\documentclass[10pt,twocolumn,letterpaper]{article}

\usepackage{cvpr}
\usepackage{times}
\usepackage{epsfig}
\usepackage{graphicx}
\usepackage{amsmath}
\usepackage{amssymb}
\usepackage{comment}
\usepackage{multirow}
\usepackage{float}
\usepackage{tikz}
\usepackage{calc}
\usepackage{adjustbox}
\usepackage{pdfpages}
\usepackage{comment}

\definecolor{newcolor}{rgb}{0,0.4,0}
\definecolor{oldcolor}{rgb}{0.7,0.7,0.7}

\ifdefined\SHOWDIFF
  \includecomment{NEW}
  \includecomment{OLD}
  \specialcomment{NEW}{\begingroup\color{newcolor}}{\endgroup}
  \specialcomment{OLD}{\begingroup\color{oldcolor}}{\endgroup}
\else
  \includecomment{NEW}
  \excludecomment{OLD}
\fi

\IfFileExists{../images/teaser.pdf}{%
\graphicspath{{../images/}}
}
{
\graphicspath{{./images/}}
}

\usepackage[pagebackref=true,breaklinks=true,letterpaper=true,colorlinks,bookmarks=false]{hyperref}

\cvprfinalcopy % *** Uncomment this line for the final submission

\def\cvprPaperID{2080} % *** Enter the CVPR Paper ID here

\ifcvprfinal\fi
\begin{document}
\newlength{\capvspace}
\setlength{\capvspace}{-0.7em}

\newlength{\figvspace}
\setlength{\figvspace}{-0.5em}

\definecolor{dgreen}{rgb}{0,.7,0}
\definecolor{ben}{rgb}{0.9,.4,0}
\definecolor{dred}{rgb}{.7,0,0}
\definecolor{dblue}{rgb}{0,0,0.7}
\definecolor{alexey}{rgb}{0.7,0,1}
\newcommand{\TODO}[1]{{\color{dblue}[TODO #1]}}

\newcommand{\tcb}[1]{\textcolor{dred}{\scriptsize #1}}    % Thomas's comments
\newcommand{\tca}[1]{\textcolor{alexey}{\scriptsize #1}}    % Alexey's comments
\newcommand{\tcu}[1]{\textcolor{ben}{\scriptsize #1}}    % Benjamin's comments
\newcommand{\tch}[1]{\textcolor{dgreen}{\scriptsize #1}}    % Huizhong's comments
\newcommand{\tcj}[1]{\textcolor{dblue}{\scriptsize #1}}    % Jonas' comments

% math commands
\newcommand{\vect}[1]{\mathbf{#1}}
\newcommand{\img}{I}
\newcommand{\depth}{D}
\newcommand{\motion}{M}
\newcommand{\preddepth}{\widehat{D}}
\newcommand{\predmotion}{\widehat{M}}
\newcommand{\weights}{w}
\newcommand{\loss}{\mathcal{L}}

\newcommand{\no}{\textcolor{dred}{no}}
\newcommand{\yes}{\textcolor{dgreen}{yes}}

\newcommand{\fig}[1]{\mbox{Fig.~\ref{#1}}}
\newcommand{\tab}[1]{\mbox{Tab.~\ref{#1}}}

\newcommand{\bb}[1]{\textbf{#1}}

\newcommand{\compw}{0.09}
\newcommand{\compc}{\textcolor{dred}}

% includegraphics with annotation/title
\newcommand{\incgraphics}[3]{%
  \begin{tikzpicture}%
    \node[inner sep=0pt] (mynode) {\adjustimage{#1}{#2}};%
    \node at (mynode) [inner sep=0,scale=1] {\strut #3};%
    %\draw [blue] (current bounding box.south west) rectangle (current bounding box.north east);%
  \end{tikzpicture}%
}
\newcommand{\incgraphicsbelow}[3]{%
  \begin{tikzpicture}%
    \node[inner sep=0pt] (mynode) {\adjustimage{#1}{#2}};%
    \node at (mynode.south) [inner sep=0,scale=1,below=1mm] {\strut #3};%
    %\draw [blue] (current bounding box.south west) rectangle (current bounding box.north east);%
  \end{tikzpicture}%
}
\newcommand{\incgraphicsabove}[4][1mm]{%
  \begin{tikzpicture}%
    \node[inner sep=0pt] (mynode) {\adjustimage{#2}{#3}};%
    \node at (mynode.north) [inner sep=0,scale=1,above=#1] {\strut #4};%
    %\draw [blue] (current bounding box.south west) rectangle (current bounding box.north east);%
  \end{tikzpicture}%
}

\newcommand{\incgraphicsbrbox}[3]{%
  \begin{tikzpicture}%
    \node[inner sep=0pt] (mynode) {\adjustimage{#1}{#2}};%
    \node at (mynode.south east) [inner sep=0.25mm,scale=1,above left=0.15ex, fill=white, opacity=0.5, text opacity=1] {#3};%
    %\draw [blue] (current bounding box.south west) rectangle (current bounding box.north east);%
  \end{tikzpicture}%
}

%%%%%%%%% TITLE
\title{DeMoN: Depth and Motion Network for Learning Monocular Stereo}

\author{Benjamin Ummenhofer\textsuperscript{*,1} \qquad Huizhong Zhou\textsuperscript{*,1}\\
%University of Freiburg\\
{\tt\small \{ummenhof, zhouh\}@cs.uni-freiburg.de}
\and
Jonas Uhrig\textsuperscript{1,2}\quad Nikolaus Mayer\textsuperscript{1}\quad Eddy Ilg\textsuperscript{1}\quad Alexey Dosovitskiy\textsuperscript{1}\quad Thomas Brox\textsuperscript{1}\\[1mm]
\textsuperscript{1}University of Freiburg \qquad \textsuperscript{2}Daimler AG R\&D\\
{\tt\small \{uhrigj, mayern, ilg, dosovits, brox\}@cs.uni-freiburg.de}
}

\maketitle
\renewcommand*{\thefootnote}{\fnsymbol{footnote}}
\footnotetext[1]{Equal contribution}
% do not allow figures in left column on first page
\global\csname @topnum\endcsname 0
\global\csname @botnum\endcsname 0

%%%%%%%%% ABSTRACT
\begin{abstract}
In this paper we formulate structure from motion as a learning problem.
We train a convolutional network end-to-end to compute depth and camera motion from successive, unconstrained image pairs.
The architecture is composed of multiple stacked encoder-decoder networks, the core part being an iterative network that is able to improve its own predictions.
The network estimates not only depth and motion, but additionally surface normals, optical flow between the images and confidence of the matching.
A crucial component of the approach is a training loss based on spatial relative differences.
Compared to traditional two-frame structure from motion methods, results are more accurate and more robust.
In contrast to the popular depth-from-single-image networks, DeMoN learns the concept of matching and, thus, better generalizes to structures not seen during training.
\end{abstract}

%%%%%%%%% BODY TEXT
\section{Introduction}

\begin{figure}
\begin{center}
  \includegraphics[width=0.43\textwidth]{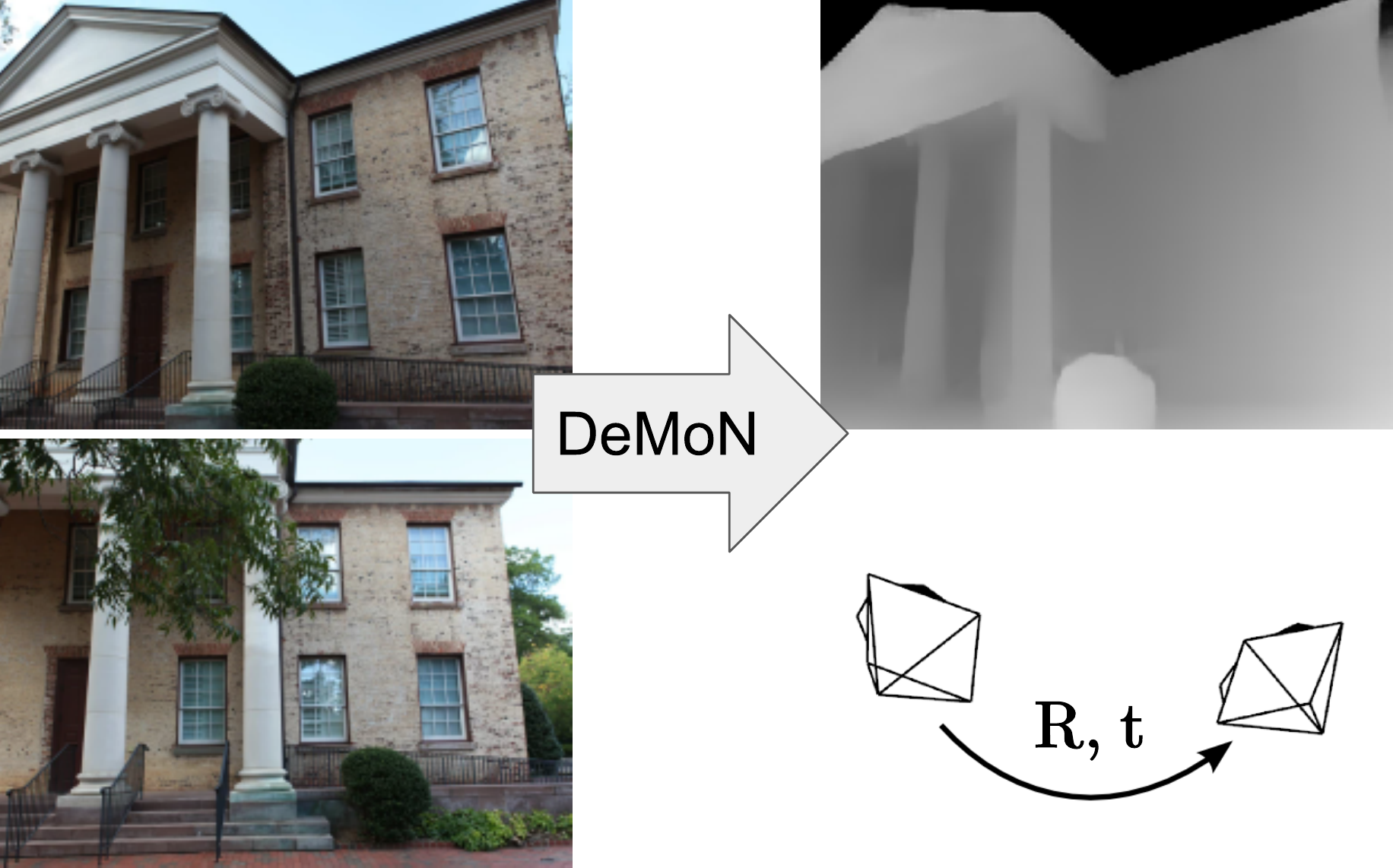}
\end{center}
\vspace{\capvspace}%
\caption{%
Illustration of DeMoN. The input to the network is two successive images from a monocular camera. The network estimates the depth in the first image and the camera motion.
}
\label{fig:teaser}
\vspace{\figvspace}
\end{figure}

%%%%% Structure for introduction

Structure from motion (SfM) is a long standing task in computer vision. 
% Sophisticated systems have been developed that are quite mature, set the state of the art, and some are even part of commercial products.
Most existing systems, which represent the state of the art, are carefully engineered pipelines consisting of several consecutive processing steps.
A fundamental building block of these pipelines is the computation of the structure and motion for two images. 
Present implementations of this step have some inherent limitations. %do not leverage all available information.
For instance, it is common to start with the estimation of the camera motion before inferring the structure of the scene by dense correspondence search. 
Thus, an incorrect estimate of the camera motion leads to wrong depth predictions. 
Moreover, the camera motion is estimated from sparse correspondences computed via keypoint detection and descriptor matching. 
This low-level process is prone to outliers and does not work in non-textured regions. 
Finally, all existing SfM approaches fail in case of small camera translation. 
This is because it is hard to integrate priors that could provide reasonable solutions in those degenerate cases.

In this paper, we succeed for the first time in training a convolutional network to jointly estimate the depth and the camera motion from an unconstrained pair of images. 
This approach is very different from the typical SfM pipeline in that it solves the problems of motion and dense depth estimation jointly.
We cannot yet provide a full learning-based system for large-scale SfM, but the two-frame case is a crucial first step towards this goal. 
% \tca{I removed the sentence about multi-frame here. What about selling it this way: single-frame is easy, Eigen et al. and many others did it; two-frame is much more difficult and we now did it; multiframe is still to be done.}
% \tcb{This would be another possibility, but then we must start with single-image, which lets our approach look like just an extension of Eigen et al. On the other hand, we must mention the limitation to two frames and that multiple frames are still missing. So far, I don't see a better version than adding my original sentence again.}
In the longer term, the learning approach has large potential, since it integrates naturally all the shape from X approaches: multi-view, silhouettes, texture, shading, defocus, haze. 
Moreover, strong priors on objects and structure can be learned efficiently from data and regularize the problem in degenerate cases; see \fig{fig:small_baseline} for example. 
This potential is indicated by our results for the two-frame scenario, where the learning approach clearly outperforms traditional methods. 
%Finally, connecting our network with a network for 3D recognition should be much easier than with conventional structure from motion. 

Convolutional networks recently have shown to excel at depth prediction from a single image \cite{eigen_predicting_2015,eigen_predicting_2014,liu_learning_2015}.
By learning priors about objects and their shapes these networks reach remarkably good performance in restricted evaluation scenarios such as indoor or driving scenes. 
However, single-image methods have more problems generalizing to previously unseen types of images.
This is because they do not exploit stereopsis. 
\fig{fig:generalization} shows an example, where depth from a single image fails, because the network did not see similar structures before. 
Our network, which learned to exploit the motion parallax, does not have this restriction and generalizes well to very new scenarios. 

To exploit the motion parallax, the network must put the two input images in correspondence.
%- a learning task that is harder to solve than depth from single image. 
We found that a simple encoder-decoder network fails to make use of stereo: when trained to compute depth from two images it ends up using only one of them.
Depth from a single image is a shortcut to satisfy the training objective
% . It seems to be a simpler learning task for the network than 
without putting the two images into correspondence and deriving camera motion and depth from these correspondences.

In this paper, we present a way to avoid this shortcut and elaborate on it to obtain accurate depth maps and camera motion estimates. The key to the problem is an architecture that alternates optical flow estimation with the estimation of camera motion and depth; see \fig{fig:encoder_decoder_pair}. 
In order to solve for optical flow, the network \emph{must} use both images. To this end, we adapted the FlowNet architecture \cite{dosovitskiy_flownet_2015} to our case. 
Our network architecture has an iterative part that is comparable to a recurrent network, since weights are shared. 
Instead of the typical unrolling, which is common practice when training recurrent networks, we append predictions of previous training iterations to the current minibatch.
This training technique saves much memory and allows us to include more iterations for training. 
%This strategy is further supported by a predictor for the confidence of the network. 
Another technical contribution of this paper is a special gradient loss to deal with the scale ambiguity in structure from motion. The network was trained on a mixture of real images from a Kinect camera, including the SUN3D dataset \cite{xiao_sun3d_2013}, and a variety of rendered scenes that we created for this work.

\section{Related Work}

Estimation of depth and motion from pairs of images goes back to Longuet-Higgins~\cite{longuet-higgins_computer_1981}. The underlying 3D geometry is a consolidated field, which is well covered in textbooks \cite{Hartley2004,Faugeras1993}.
State-of-the-art systems~\cite{frahm_building_2010,wu_towards_2013} allow for reconstructions of large scenes including whole cities. They consist of a long pipeline of methods, starting with 
%a procedure to calibrate the internal parameters of the camera \cite{pollefeys_stratified_1997}, 
descriptor matching for finding a sparse set of correspondences between images~\cite{lowe_distinctive_2004}, followed by estimating the essential matrix to determine the camera motion. Outliers among the correspondences are typically filtered out via RANSAC \cite{fischler_random_1981}.
Although these systems use bundle adjustment \cite{triggs_bundle_2000} to jointly optimize camera poses and structure of many images, they 
depend on the quality of the estimated geometry between image pairs for initialization.
Only after estimation of the camera motion and a sparse 3D point cloud, dense depth maps are computed by exploiting the epipolar geometry \cite{collins_space-sweep_1996}. 
LSD-SLAM \cite{engel14eccv} deviates from this approach by jointly optimizing semi-dense correspondences and depth maps. It considers multiple frames from a short temporal window but does not include bundle adjustment.
DTAM \cite{newcombe_dtam:_2011} can track camera poses reliably for critical motions by matching against dense depth maps.
However, an external depth map initialization is required, which in turn relies on classic structure and motion methods.

Camera motion estimation from dense correspondences has been proposed by Valgaerts et al.~\cite{valgaerts_dense_2012}. 
In this paper, we deviate completely from these previous approaches by training a single deep network that includes computation of dense correspondences, estimation of depth, and the camera motion between two frames. 

% single image
%Convolutional networks (ConvNets) are mainly being used for recognition, but more and more other computer vision tasks were being formulated as deep learning problems, too. 
Eigen et al.~\cite{eigen_predicting_2015} trained a ConvNet to predict depth from a single image.
%on the NYU dataset of indoor videos recorded with a Kinect camera~\cite{NYU}.
Depth prediction from a single image is an inherently ill-posed problem which can only be solved using priors and semantic understanding of the scene~-- tasks ConvNets are known to be very good at.
Liu et al.~\cite{liu_learning_2015} combined a ConvNet with a superpixel-based conditional random field, yielding improved results.
Our two-frame network also learns to exploit the same cues and priors as the single-frame networks, but in addition it makes use of a pair of images and the motion parallax between those. 
This enables generalization to arbitrary new scenes. 
% Nonetheless, our two-frame network learns to exploit the same cues and priors as the single-frame networks. 
% Thus, our contribution is complementary to single-frame networks.

ConvNets have been trained to replace the descriptor matching module in aforementioned SfM systems~\cite{Exemplar_CNN_PAMI, zagoruyko_2015}.
%showed that features from AlexNet~\cite{krizhevsky_alexnet_2012}, a large network trained for classification on the ImageNet dataset, performs well on a descriptor matching task, and features learned in unsupervised fashion perform yet better.
%Zagoruyko et al.~\cite{zagoruyko_2015} trained a network specifically for the descriptor matching task, reaching very good results.
The same idea was used by {\v{Z}}bontar and LeCun~\cite{zbontar_computing_2015} to estimate dense disparity maps between stereo images. 
% The actual optimization on the disparity maps is left to classical  semi-global matching~(SGM)~\cite{hirschmuller_accurate_2005}.
Computation of dense correspondences with a ConvNet that is trained end-to-end on the task, was presented by Dosovitskiy et al.~\cite{dosovitskiy_flownet_2015}. Mayer et al.~\cite{mayer_sceneflownet_2016} applied the same concept to dense disparity estimation in stereo pairs.
We, too, make use of the FlowNet idea \cite{dosovitskiy_flownet_2015}, but in contrast to \cite{mayer_sceneflownet_2016,zbontar_computing_2015}, the motion between the two views is not fixed, but must be estimated to derive depth estimates.
This makes the learning problem much more difficult. 
\begin{figure*}
\begin{center}
\includegraphics[width=1.0\textwidth]{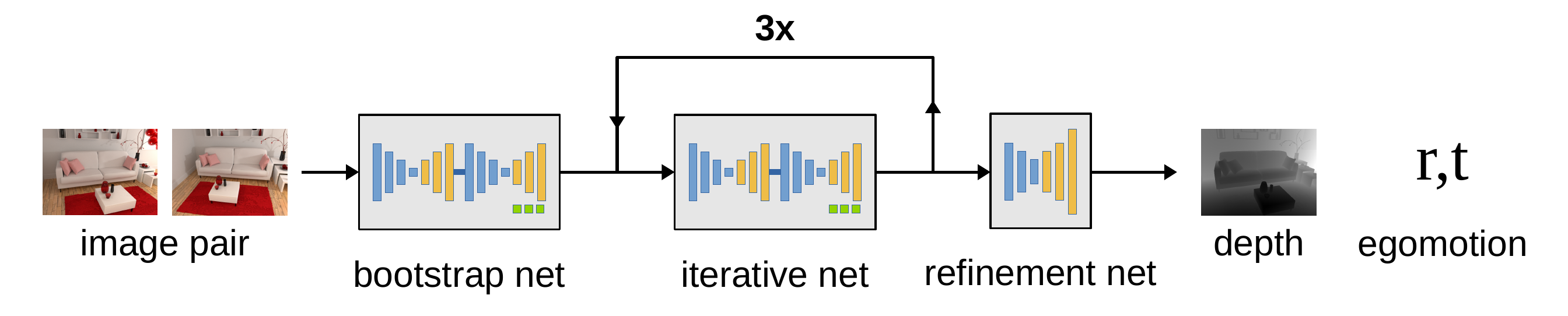}
\end{center}
\vspace{-0.9em}%
\vspace{\capvspace}%
\caption{%
Overview of the architecture. DeMoN takes an image pair as input and predicts the depth map of the first image and the relative pose of the second camera. The network consists of a chain of encoder-decoder networks that iterate over optical flow, depth, and egomotion estimation; see \fig{fig:encoder_decoder_pair} for details. The refinement network increases the resolution of the final depth map.
%\TODO{Remove artifact in top right corner of the figure. Fixed in git} 
%\TODO{this figure would look nicer on page 2} \tcb{Fig.2 and Fig.3 must stay together.}
%The bootstrap and iterative networks consist of pairs of encoder-decoder networks.
%In both cases the first encoder predicts optical flow and its confidence and the second encoder-decoder predicts depth and camera motion.
%Each encoder-decoder uses predictions from its predecessor.
%The refinement network uses a single encoder to refine and upsample the depth prediction.
}
\label{fig:network_overview}
\end{figure*}

\begin{figure*}
\begin{center}
\includegraphics[page=1,width=\textwidth]{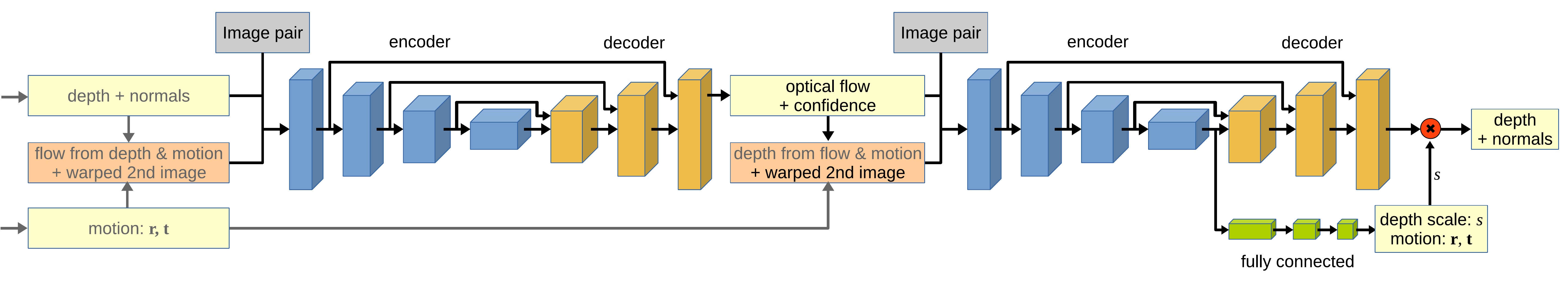}
\end{center}
\vspace{-0.8em}%
\vspace{\capvspace}%
\caption{%
Schematic representation of the encoder-decoder pair used in the bootstrapping and iterative network.
Inputs with gray font are only available for the iterative network.
The first encoder-decoder predicts optical flow and its confidence from an image pair and previous estimates.
The second encoder-decoder predicts the depth map and surface normals. A fully connected network appended to the encoder estimates the camera motion $\vect r, \vect t$ and a depth scale factor $s$.
The scale factor $s$ relates the scale of the depth values to the camera motion.
%We use this factor to scale the final depth values.
%\TODO{Where do those blue dots next to the layers come from?!}\tcu{probably old libreoffice version. I will fix it on my notebook}
}
\label{fig:encoder_decoder_pair}
\vspace{\figvspace}
\end{figure*}

%% depth from video
Flynn et al.~\cite{flynn_deepstereo_2015} implicitly estimated the 3D structure of a scene from a monocular video using a convolutional network.
They assume known camera poses~-- a large simplification which allows them to use the plane-sweeping approach to interpolate between given views of the scene.
Moreover, they never explicitly predict the depth, only RGB images from intermediate viewpoints.

%% camera pose
% motion
Agrawal et al.~\cite{agrawal_egomotion_2015} and Jayaraman \& Grauman~\cite{Jayaraman2015egomotion} applied ConvNets to estimating camera motion. The main focus of these works is not on the camera motion itself, but on learning a feature representation useful for recognition. The accuracy of the estimated camera motion is not competitive with classic methods. 
%They were only using the camera motion for feature learning and therefore did not compare the performance with any baseline methods.
% relocalization
Kendall et al.~\cite{kendall_modelling_2015} trained a ConvNet for camera relocalization~--- predicting the location of the camera within a known scene from a single image.
%To this end they trained the network in supervised manner on frames from this scene.
This is mainly an instance recognition task and requires retraining for each new scene. 
All these works do not provide depth estimates.

%%%%%%%%%%%%%%%%%%%%%%%%%%%%%%%%%%%%%%%
\section{Network Architecture}

The overall network architecture is shown in \fig{fig:network_overview}.
\mbox{DeMoN} is a chain of encoder-decoder networks solving different tasks. 
The architecture consists of three main components: the bootstrap net, the iterative net and the refinement net. 
%Individual design decisions are justified in Section~\ref{sec:ablation_study}.
The first two components are pairs of encoder-decoder networks, where the first one computes optical flow while the second one computes depth and camera motion; see \fig{fig:encoder_decoder_pair}. The iterative net is applied 
% multiple times with the same weights 
recursively to successively refine the estimates of the previous iteration. 
The last component is a single encoder-decoder network that generates the final upsampled and refined depth map.% and camera motion.

\bb{Bootstrap net.}
The bootstrap component gets the image pair as input and outputs the initial depth and motion estimates. 
Internally, first an encoder-decoder network computes optical flow and a confidence map for the flow (the left part of \fig{fig:encoder_decoder_pair}). 
% The left part of  shows the structure of this encoder-decoder.
The encoder consists of pairs of convolutional layers with 1D filters in $y$ and $x$-direction.
Using pairs of 1D filters as suggested in \cite{szegedy_rethinking_2015} allows us to use spatially large filter while keeping the number of parameters and runtime manageable.
We gradually reduce the spatial resolution with a stride of $2$ while increasing the number of channels.
The decoder part generates the optical flow estimate from the encoder's representation via a series of up-convolutional layers with stride $2$ followed by two convolutional layers.
% to reduce the number of channels to $4$.
It outputs two components of the optical flow field and an estimate of their confidence.
Details on the loss and the training procedure are described in Section~\ref{sec:training}.
%We generate the optical flow and the confidence at a spatial resolution of $64\times48$ and $16\times12$.
%The smaller prediction is only used during training.
%The resolution of the input images is $256\times192$.

The second encoder-decoder, shown in the right part of \fig{fig:encoder_decoder_pair}, takes as input the optical flow, its confidence, the image pair, and the 
% warped second image that compensates for the estimated optical flow.
second image warped with the estimated flow field.
Based on these inputs it estimates depth, surface normals, and camera motion.
The architecture is the same as above, apart from the extra 3 fully connected layers that compute the camera motion and a scaling factor for the depth prediction.
The latter reflects the inherent connection between depth and motion predictions due to scale ambiguity; see Section~\ref{sec:parametrization}. 

\begin{table}
  \begin{center}
    \setlength{\tabcolsep}{0.1cm}
    { \small
    \begin{tabular}{|c|ccc|}
      \hline
       Method              &  L1-inv &  sc-inv &  L1-rel \\
      \hline
      Single image         &   0.080 &   0.159 &   0.696 \\
      Na{\"i}ve image pair &   0.079 &   0.165 &   0.722 \\
      DeMoN                &   \bb{0.012} & \bb{0.131} & \bb{0.097} \\
      
    %   single-kinect &   0.111 &   0.174 &   0.198 \\
    %   pair-kinect   &   0.113 &   0.179 &   0.201 \\
    \hline
    \end{tabular}
    }
  \end{center}
  \vspace{0.4em}%
  \vspace{\capvspace}%
  \caption{%
Na{\"i}ve two-frame depth estimation does not perform better than depth from a single image on any of the error measures (smaller is better). 
% A single image is sufficient to bring the training error down. 
The architecture of DeMoN forces the network to use both images, yielding a large performance improvement. 
% A network using both images would be able to estimate the scale dependent inverse depth more accurately, because it can infer the scale from the second image (assuming a normalized translation of the camera). 
}
  \label{tbl:ablation_naive_two_frame}
  \vspace{\figvspace}
  \vspace{-0.2em}
\end{table}

% Thanks to the optical flow input, there are some rough dense point correspondences that keep this network from just using one of the two input frames. 
By feeding optical flow estimate into the second encoder-decoder we let it make use of motion parallax.
\tab{tbl:ablation_naive_two_frame} shows that an encoder-decoder network trained to estimate depth and camera motion  directly from an image pair (na{\"i}ve image pair) fails to make use of stereo cues and performs on par with a single-image network.  
DeMoN, on the other hand, performs significantly better.
%The similarity of the warped second image to the first image allows to assess the quality of the optical flow prediction.

\bb{Iterative net.}
The iterative net is trained to improve existing depth, normal, and motion estimates.
The architecture of this encoder-decoder pair is identical to the bootstrap net, but it takes additional inputs. 
We convert the depth map and camera motion estimated by the bootstrap net or a previous iteration of the iterative net into an optical flow field, and feed it into the first encoder-decoder together with other inputs.
Likewise, we convert the optical flow to a depth map using the previous camera motion prediction and pass it along with the optical flow to the second encoder-decoder.
In both cases the networks are presented with a prediction proposal generated from the predictions of the previous encoder-decoder.
%Note that the camera motion used to convert the optical flow to a depth map is an exception we make, to avoid computing the camera motion directly from optical flow, which may fail.

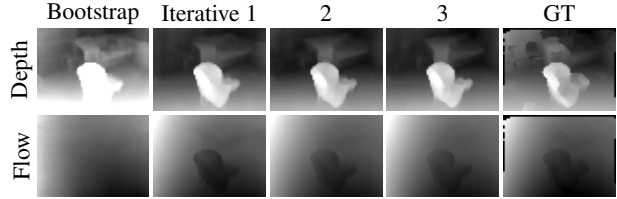
\begin{figure}
\begin{center}
  \begin{tikzpicture}[scale=.25]
    \node at (5.0,3) {\small Bootstrap};
    \node at (11.2,3) {\small Iterative 1};
    \node at (17.4,3) {\small 2};
    \node at (23.6,3) {\small 3};
    \node at (29.8,3) {\small GT};
    \node at (1.2,0) [rotate=90] {\small Depth};
    \node at (5,0) {\includegraphics[width=0.085\textwidth]{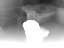}};
    \node at (11.2,0) {\includegraphics[width=0.085\textwidth]{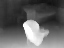}};
    \node at (17.4,0) {\includegraphics[width=0.085\textwidth]{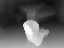}};
    \node at (23.6,0) {\includegraphics[width=0.085\textwidth]{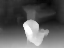}};
    \node at (29.8,0) {\includegraphics[width=0.085\textwidth]{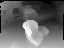}};
    \node at (1.2,-4.6) [rotate=90] {\small Flow};
    \node at (5.0,-4.6) {\includegraphics[width=0.085\textwidth]{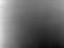}};
    \node at (11.2,-4.6) {\includegraphics[width=0.085\textwidth]{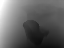}};
    \node at (17.4,-4.6) {\includegraphics[width=0.085\textwidth]{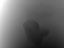}};
    \node at (23.6,-4.6) {\includegraphics[width=0.085\textwidth]{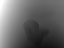}};
    \node at (29.8,-4.6) {\includegraphics[width=0.085\textwidth]{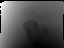}};
  \end{tikzpicture}
\end{center}
\vspace{\capvspace}%
\caption{
\bb{Top:} Iterative depth refinement. 
The bootstrap net fails to accurately estimate the scale of the depth.
The iterations refine the depth prediction and strongly improve the scale of the depth values. 
The L1-inverse error drops from $0.0137$ to $0.0072$ after the first iteration.  
\bb{Bottom:} Iterative refinement of optical flow.
Images show the $x$ component of the optical flow for better visibility. 
%\tcb{Why not showing the full optical flow as color plot? Easier to recognize it as flow. Edges less visible?}
The flow prediction of the bootstrap net misses the object completely.
Motion edges are retrieved already in the first iteration and the endpoint error is reduced from 0.0176 to 0.0120.
 %\tch{epe decreases and then increases unfortunately}$0.0199$, $0.0122$, $0.0133$, $0.0130$.
}
\label{fig:predictions_iterations}
\vspace{\figvspace}
\end{figure}

%Input and output are the depth map and the surface normals for the first camera, and the relative pose of the second camera, which makes it possible to run this group in an iterative manner.
\fig{fig:predictions_iterations} shows how the optical flow and depth improve with each iteration of the network. 
The iterations enable sharp discontinuities, improve the scale of the depth values, and can even correct wrong estimates of the initial bootstrapping network.
%\tcu{on average this is true, there are outliers. can we find images which visually show these effects?}
The improvements largely saturate after 3 or 4 iterations.
Quantitative analysis is shown in the supplementary material.

During training we simulate 4 iterations by appending predictions from previous training iterations to the minibatch.
Unlike unrolling, there is no backpropagation of the gradient through iterations.
Instead, the gradient of each iteration is described by the losses on the well defined network outputs:  optical flow, depth, normals, and camera motion.
Compared to backpropagation through time this saves a lot of memory and allows us to have a larger network and more iterations. 
\begin{NEW}
A similar approach was taken by Li~\etal~\cite{li_iterative_2016}, who train each iteration in a separate step and therefore need to store predictions as input for the next stage.
We also train the first iteration on its own, but then train all iterations jointly which avoids intermediate storage.
\end{NEW}

\begin{figure}
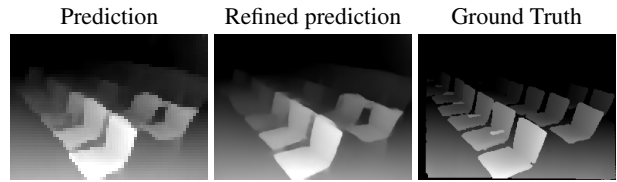

\begin{center}
  \incgraphicsabove[0mm]{width=0.15\textwidth}{png/refine/42_iter3_predicted_depth.png}{\small Prediction}
  \incgraphicsabove[0mm]{width=0.15\textwidth}{png/refine/42_iter3_refined_predicted_depth.png}{\small Refined prediction}
  \incgraphicsabove[0mm]{width=0.15\textwidth}{png/refine/42_depth.png}{\small Ground Truth}
\end{center}
\vspace{\capvspace}%
\caption{%
The refinement net generates a high-resolution depth map ($256\times192$) from the low-resolution estimate ($64\times48$) and the input image.
The depth sampling preserves depth edges and can even repair wrong depth measurements. %\TODO{put sculpture image with repaired hole here.. Jonas: no, that one has too much (very!) bad background. Any other good examples?}
}
\label{fig:refinement_comparison}
\vspace{\figvspace}
\end{figure}

\bb{Refinement net.}
% single encoder-decoder
% show figure with before/after refinement comparison + numbers
While the previous network components operate at a reduced resolution of $64\times48$ to save parameters and to reduce training and test time, the final refinement net upscales the predictions to the full input image resolution ($256\times192$).  
It gets as input the full resolution first image and the nearest-neighbor-upsampled depth and normal field.
%\TODO{is it a bilinearly upsampled version or nearest neighbor? eddy says nearest neighbor works better for flow!}
\fig{fig:refinement_comparison} shows the low-resolution input and the refined high-resolution output.

A forward pass through the network with 3 iterations takes 110ms on an Nvidia GTX Titan X. 
Implementation details and exact network definitions of all network components are provided in the supplementary material.

%%%%%%%%%%%%%%%%%%%%%%%%%%%
\section{Depth and Motion Parameterization}
\label{sec:parametrization}
% inverse depth, normalized translation, rotation as scaled axis angle
% describe scale ambiguity
% inverse depth, normalized translation, rotation as scaled axis angle
% describe scale ambiguity
The network computes the depth map in the first view and the camera motion to the second view.
We represent the relative pose of the second camera with $\vect r, \vect t \in \mathbb R^3$.
The rotation $\vect r = \theta \vect v$ is an angle axis representation with angle $\theta$ and axis $\vect v$.
The translation $\vect t$ is given in Cartesian coordinates.

It is well-known that the reconstruction of a scene from images with unknown camera motion can be determined only up to scale.
%Let $\vect X$ be a point in the world coordinate frame and $\vect x=(x,y,z)^\top$ be the image point in homogeneous coordinates.
%Given the calibration matrix $\vect K$, the rotation $\vect R$ as $3\times3$ matrix and the translation $\vect t$, we can compute the image point as $\vect x = \vect K(\vect R \alpha \vect X + \alpha \vect t)$ with an unknown scale factor $\alpha \neq 0$.
%It is obvious that scaling has no effect on the inhomogeneous image point which is $(x/z,y/z)^\top$.
We resolve the scale ambiguity by normalizing translations and depth values such that $\Vert \vect t \Vert = 1$.
This way the network learns to predict a unit norm translation vector.

%\tcu{here it becomes very obvious, that having no camera motion would most likely break the prediction - what can we do?}

Rather than the depth $z$, the network estimates the inverse depth $\xi = 1/z$.
%, where $z$ is the distance along the optical axis in the coordinate frame of the first camera.
The inverse depth allows representation of points at infinity and accounts for the growing localization uncertainty of points with increasing distance.% with a smaller penalty.
To match the unit translation, our network predicts a scalar scaling factor $s$, which we use to obtain the final depth values $s\xi$.

\section{Training Procedure}
\label{sec:training}

\subsection{Loss functions}
% Losses absolute, gradient, flow, depthtoflow

The network estimates outputs of very different nature: high-dimensional (per-pixel) depth maps and low-dimensional camera motion vectors. 
%Thus, definition of a good loss function is non-trivial. 
% The loss has to be designed in such a way that none of these objectives gets suppressed by the other, and ideally they even support each other.
The loss has to balance both of these objectives, and stimulate synergy of the two tasks without over-fitting to a specific scenario. 
% We performed multiple experiments with different loss functions: separately comparing the depth and the motion predictions, as well as losses involving both.

\bb{Point-wise losses.}
We apply point-wise losses to our outputs: inverse depth $\xi$, surface normals $\vect n$, optical flow $\vect w$, and optical flow confidence $\vect c$.
For depth we use an L1 loss directly on the inverse depth values:
%, which emphasizes the importance of close objects.
% The point-wise loss for the inverse depth is
\begin{equation}
\label{eq:loss_depth}
%\loss_\text{depth} = \sum_{i,j} \left\vert s\xi(i,j) - \hat{\xi}(i,j) \right\vert, 
\loss_\text{depth} = \textstyle{\sum_{i,j}} \vert s\xi(i,j) - \hat{\xi}(i,j) \vert, 
\end{equation}
with ground truth $\hat\xi$.
Note that we apply the predicted scale $s$ to the predicted values $\xi$.

For the loss function of the normals and the optical flow we use the (non-squared) L2 norm to penalize deviations from the respective ground truths $\hat{\vect n}$ and $\hat{\vect w}$
\begin{equation}
\begin{aligned}
%\loss_\text{normal} &= \sum_{i,j} \left\Vert \vect n(i,j) - \hat{\vect n}(i,j) \right\Vert_2 \\
%\loss_\text{flow} &= \sum_{i,j} \left\Vert \vect w(i,j) - \hat{\vect w}(i,j) \right\Vert_2.
\loss_\text{normal} &= \textstyle{\sum_{i,j}} \left\Vert \vect n(i,j) - \hat{\vect n}(i,j) \right\Vert_2 \\
\loss_\text{flow} &= \textstyle{\sum_{i,j}} \left\Vert \vect w(i,j) - \hat{\vect w}(i,j) \right\Vert_2.
\end{aligned}
\end{equation}
For optical flow this amounts to the usual endpoint error.
%The depth loss $\loss_\text{depth}$ acts on the inverse depth values which emphasizes the error for close objects.
%For the normal loss $\loss_\text{normal}$ we use the L2 norm, which penalizes deviations in the Euclidean space.

%Flow confidence ground truth for the $x$ component is defined as $\hat{c}_x(i,j) $ and then point-wise L1 loss is applied:
%\begin{equation}
%\begin{aligned}
%\hat{c}_x(i,j) &= e^{- \vert\vect w_x(i,j) - \hat{\vect w}_x(i,j)\vert} \\
%\loss_{\text{flow confidence}\; x} &= \sum_{i,j} \left\vert c_x(i,j) - \hat{c}_x(i,j) \right\vert.
%\end{aligned}
%\end{equation}

 We train the network to assess the quality of its own flow prediction by predicting a confidence map for each optical flow component.
 The ground truth of the confidence for the x component is 
 \begin{equation}
 \hat{c}_x(i,j) = e^{- \vert\vect w_x(i,j) - \hat{\vect w}_x(i,j)\vert},
 \end{equation}
 and the corresponding loss function reads as
 \begin{equation}
 %\loss_\text{flow confidence} = \sum_{i,j} \left\vert c_x(i,j) - \hat{c}_x(i,j) \right\vert.
 \loss_\text{flow confidence} = \textstyle{\sum_{i,j}} \left\vert c_x(i,j) - \hat{c}_x(i,j) \right\vert.
 \end{equation}

\bb{Motion losses.}
We use a minimal parameterization of the camera motion with 3 parameters for rotation $\vect r$ and translation $\vect t$ each.
The losses for the motion vectors are
\begin{equation}
\begin{aligned}
\loss_\text{rotation} &= \Vert \vect r - \hat{\vect r} \Vert_2 \\
\loss_\text{translation} &= \Vert \vect t - \hat{\vect t} \Vert_2.
\end{aligned}
\end{equation}
The translation ground truth is always normalized such that $\Vert \hat{\vect t} \Vert_2 = 1$, while the magnitude of $\hat{\vect r}$ encodes the angle of the rotation.

\bb{Scale invariant gradient loss.}
We define a discrete scale invariant gradient $\vect g$ as
\begin{equation}
\vect g_h[f](i,j) = \left( \tfrac{f(i+h,j) - f(i,j)}{\vert f(i+h,j) \vert + \vert f(i,j) \vert}, \tfrac{f(i,j+h) - f(i,j)}{\vert f(i,j+h) \vert + \vert f(i,j) \vert} \right)^\top.
\end{equation}
Based on this gradient we define a scale invariant loss that penalizes relative depth errors between neighbouring pixels:
\begin{equation}
\label{eq:loss_grad}
\loss_{\text{grad}\, \xi} = \sum_{h \in \{1,2,4,8,16\}} \sum_{i,j} \left\Vert \vect g_h[\xi](i,j) - \vect g_h[\hat\xi](i,j)\right\Vert_2.
\end{equation}
To cover gradients at different scales we use 5 different spacings $h$.
This loss stimulates the network to compare depth values within a local neighbourhood for each pixel.
It emphasizes depth discontinuities, stimulates sharp edges in the depth map and increases smoothness within homogeneous regions as seen in \fig{fig:gradient_comparison}.
Note that due to the relation $\vect g_h[\xi](i,j) = - \vect g_h[z](i,j)$ for $\xi, z > 0$, the loss is the same for the actual non-inverse depth values $z$.

We apply the same scale invariant gradient loss to each component of the optical flow.
% \begin{equation}
% \loss_{\text{grad} \vect w} = \sum_{h \in \{1,2,4,8,16\}} \sum_{i,j} \left\Vert \vect g_h[w_x](i,j) - \vect g_h[\widehat w_x](i,j)\right\Vert_2.
% \end{equation}
This enhances the smoothness of estimated flow fields and the sharpness of motion discontinuities.

\bb{Weighting.} We individually weigh the losses to 
% account for the different representations to 
balance their importance.
The weight factors were determined empirically and are listed in the supplementary material.

\subsection{Training Schedule}
% this is complicated:
% train networks separately to initialize, batch trick instead of concatenation
The network training is based on the Caffe framework~\cite{jia_caffe:_2014}. We train our model from scratch with Adam~\cite{kingma_adam:_2014} using a momentum of 0.9 and a weight decay of 0.0004. 
%The multi-step learning rate policy is applied. 
The whole training procedure consists of three phases.%: bootstrap, iterative and refinement training.

First, we sequentially train the four encoder-decoder components in both bootstrap and iterative nets for 250k iterations each with a batch size of 32.
While training an encoder-decoder we keep the weights for all previous components fixed.
For encoder-decoders predicting optical flow, the scale invariant loss is applied after 10k iterations.
%we train for the first 10.000 iterations without the scale invariant loss. 

%During the bootstrap phase we sequentially train the four encoder-decoders components in both bootstrap and iterative nets for 250k iterations each, initialized with a base learning rate of 0.00015 and a batch size of 32. The weights from the previous components are fixed while the next component is training. For the flow component, the first 10k iterations are trained without gradient loss to improve stability in this first training phase. 

Second, we train only the encoder-decoder pair of the iterative net.
In this phase we append outputs from previous three training iterations to the minibatch.
%During iterative training, only the encoder-decoder pair of the iterative net is trained. 
In this phase the bootstrap net uses batches of size 8. 
The outputs of the previous three network iterations are added to the batch, which yields a total batch size of 32 for the iterative network. 
We run 1.6 million training iterations.
% and use a smaller learning rate of 0.0001. 

Finally, the refinement net is trained for 600k iterations with all other weights fixed. 
The details of the training process, including the learning rate schedules, are provided in the supplementary material.

\section{Experiments}

% Baseline implementation on sun3d, tumrgbd

\subsection{Datasets}

\bb{SUN3D} \cite{xiao_sun3d_2013} provides a diverse set of indoor images together with depth and camera pose.
The depth and camera pose on this dataset are not perfect.
Thus, we sampled image pairs from the dataset and automatically discarded pairs with a high photoconsistency error. We split the dataset so that the same scenes do not appear in both the training and the test set.

\bb{RGB-D SLAM} \cite{sturm12iros} provides high quality camera poses obtained with an external motion tracking system.
Depth maps are disturbed by measurement noise, and we use the same preprocessing as for SUN3D. We created a training and a test set. 

\bb{MVS} includes several outdoor datasets. We used the Citywall and Achteckturm datasets from \cite{fuhrmann2014mve} and the Breisach dataset \cite{UB15} for training and the datasets provided with COLMAP \cite{schoenberger2016sfm,schoenberger2016mvs} for testing. The depth maps of the reconstructed scenes are often sparse and can comprise reconstruction errors.

\bb{Scenes11} is a synthetic dataset with generated images of virtual scenes with random geometry, which provide perfect depth and motion ground truth, but lack realism. 
%Another disadvantage is the simplistic model of the camera movement, which relies on parameters sampled from Gaussian and uniform distributions.
%When training only on this data, the network might learn the underlying model by heart.

Thus, we introduce the \bb{Blendswap} dataset which is based on 150 scenes from \texttt{blendswap.com}.
The dataset provides a large variety of scenes, ranging from cartoon-like to photorealistic scenes.
The dataset contains mainly indoor scenes. We used this dataset only for training. 

\bb{NYUv2} \cite{NYU} provides depth maps for diverse indoor scenes but lacks camera pose information.
We did not train on NYU and used the same test split as in Eigen et al.~\cite{eigen_predicting_2015}.
In contrast to Eigen et al., we also require a second input image that should not be identical to the previous one. Thus, we automatically chose the next image that is sufficiently different from the first image according to a threshold on the difference image.

In all cases where the surface normals are not available, we generated them from the depth maps.
We trained DeMoN specifically for the camera intrinsics used in SUN3D and adapted all other datasets by cropping and scaling to match these parameters.

\subsection{Error metrics}
While single-image methods aim to predict depth at the actual physical scale, two-image methods typically yield the scale relative to the norm of the camera translation vector.
%\tcb{In fact, predicting some physical scale based on prior knowledge of the structural size would be valuable for two-image methods, too. And a learning approach makes this possible. }\tcu{I agree. Sounds like a good student project to me.}
Comparing the results of these two families of methods requires a scale-invariant error metric.
We adopt the scale-invariant error of \cite{eigen_predicting_2014}, which is defined as
\begin{equation}
%\text{sc-inv}(z,\hat{z}) = \sqrt{\tfrac{1}{n}\sum_i d_i^2 - \tfrac{1}{n^2} \left(\sum_i d_i\right)^2},
\textstyle \text{sc-inv}(z,\hat{z}) = \sqrt{\tfrac{1}{n}\sum_i d_i^2 - \tfrac{1}{n^2} \left(\sum_i d_i\right)^2},
\end{equation}
with $ d_i = \log z_i - \log \hat{z}_i$.
For comparison with classic structure from motion methods we use the following measures:
%\begin{align}
%\text{L1-rel}(z,\hat{z}) &= \tfrac{1}{n} \sum_i \frac{\vert z_i - \hat{z}_i\vert}{\hat{z}_i}\\
%\text{L1-inv}(z,\hat{z}) &= \tfrac{1}{n} \sum_i \vert \xi_i - \hat{\xi}_i\vert = \tfrac{1}{n} \sum_i \left\vert \frac{1}{z_i} - \frac{1}{\hat{z}_i}\right\vert 
%\end{align}
\begin{align}
\text{L1-rel}(z,\hat{z}) &= \tfrac{1}{n} \textstyle{\sum_i} \frac{\vert z_i - \hat{z}_i\vert}{\hat{z}_i}\\
\text{L1-inv}(z,\hat{z}) &= \tfrac{1}{n} \textstyle{\sum_i} \vert \xi_i - \hat{\xi}_i\vert = \tfrac{1}{n} \textstyle{\sum_i} \left\vert \frac{1}{z_i} - \frac{1}{\hat{z}_i}\right\vert 
\end{align}
L1-rel computes the depth error relative to the ground truth depth and therefore reduces errors where the ground truth depth is large and increases the importance of close objects in the ground truth.
L1-inv behaves similarly and resembles our loss function for predicted inverse depth values \eqref{eq:loss_depth}.
%For this reason when comparing the proposed approach with single-image methods we only report errors of the optimally-scaled predictions.
%That is, before computing the errors we scaled the prediction so as to minimize the MSE between the logarithms of the predicted depth and the ground truth.

For evaluating the camera motion estimation, we report the angle (in degrees) between the prediction and the ground truth for both the translation and the rotation. 
The length of the translation vector is $1$ by definition.

The accuracy of optical flow is measured by the average endpoint error (EPE), that is, the Euclidean norm of the difference between the predicted and the true flow vector, averaged over all image pixels. The flow is scaled such that the displacement by the image size corresponds to $1$.

%In this section we show that the presented approach compares favorably to a number of strong baseline methods.
%These come from two categories: methods which estimate depth and motion from an image pair (``two-frame'') and single-image depth estimation methods.
%The two-frame methods are generally applicable but cannot exploit priors which can be learned from data.
%Single-frame methods, on contrary, rely heavily on priors which do not generalize to previously unseen types of data. 
%\tca{why is Eigen so good on MVS then?}
%We show that DeMoN combines the best of both worlds.
%\tca{report average over datasets?}

\subsection{Comparison to classic structure from motion}

We compare to several strong baselines implemented by us from state-of-the-art components (``Base-*'').
For these baselines, we estimated correspondences between images, either by matching SIFT keypoints (``Base-SIFT'') or with the FlowFields  optical flow method from Bailer et al.~\cite{bailer_flow_2015} (``Base-FF'').
% We found this dense correspondence method more reliable than keypoint based methods like SIFT, as shown in the supplementary material. \tca{are we sure?}
% We found this dense correspondence method more reliable than keypoint based methods like SIFT for our data. \tcb{would work better on larger images.}
Next, we computed the essential matrix with the normalized 8-point algorithm~\cite{hartley_defense_1997} and RANSAC.
To further improve accuracy we minimized the reprojection error using the \emph{ceres} library \cite{agarwal_ceres}.
Finally, we generated the depth maps by plane sweep stereo and used the approach of Hirschmueller et al.~\cite{hirschmuller_accurate_2005} for optimization.
We also report the accuracy of the depth estimate when providing the ground truth camera motion (``Base-Oracle'').
(``Base-Matlab'') and (``Base-Mat-F'') are implemented in Matlab. (``Base-Matlab'') uses Matlab implementations of the KLT algorithm \cite{Tomasi91detectionand,lucas_iterative_1981,Shi_1994_3266} for correspondence search while (``Base-Mat-F'') uses the predicted flow from DeMoN. The essential matrix is computed with RANSAC and the 5-point algorithm \cite{nister_efficient_2004} for both.

\begin{table}
  \begin{minipage}{0.73\linewidth}
  \begin{flushleft}
    \setlength{\tabcolsep}{0.1cm}
    { \scriptsize
    \begin{tabular}{|c|c|ccc|cc|}
      \hline
      
                                                                           &             & \multicolumn{3}{c|}{Depth}             & \multicolumn{2}{c|}{Motion}    \\           
      \hline                                                               
                                                                           & Method      &  L1-inv      &  sc-inv    &  L1-rel    & rot        & trans             \\
      \hline                                                                                                                                                       
      \hline                                                                                                                                                       
      \parbox[t]{2mm}{\multirow{6}{*}{\rotatebox[origin=c]{90}{MVS}}}      & Base-Oracle & \bb{0.019}   & \bb{0.197} & \bb{0.105} &  0         &  0                \\
                                                                           & Base-SIFT   &   0.056      &   0.309    &   0.361    &  21.180    &  60.516           \\
                                                                           & Base-FF     &   0.055      &   0.308    &   0.322    & \bb{4.834} &  17.252           \\
                                                                           & Base-Matlab &     -        &     -      &     -      &  10.843    &  32.736           \\
                                                                           & Base-Mat-F  &     -        &     -      &     -      &   5.442    &  18.549
                                                                           \\                                                                           
                                                                           & DeMoN       &   0.047      &   0.202    &   0.305    &   5.156    &  \bb{14.447}      \\
      \hline                                                                                                                                                       
      \hline                                                                                                                                                       
      \parbox[t]{2mm}{\multirow{6}{*}{\rotatebox[origin=c]{90}{Scenes11}}} & Base-Oracle &   0.023      &   0.618    &   0.349    &  0         &  0                \\
                                                                           & Base-SIFT   &   0.051      &   0.900    &   1.027    &  6.179     &  56.650           \\
                                                                           & Base-FF     &   0.038      &   0.793    &   0.776    &  1.309     &  19.425           \\
                                                                           & Base-Matlab &     -        &     -      &     -      &  0.917     &  14.639           \\
                                                                           & Base-Mat-F  &     -        &     -      &     -      &  2.324     &  39.055
                                                                           \\                                                                           
                                                                           & DeMoN       &   \bb{0.019} & \bb{0.315} & \bb{0.248} & \bb{0.809} &  \bb{8.918}       \\
      \hline                                                                                                                                                       
      \hline                                                                                                                                                       
      \parbox[t]{2mm}{\multirow{6}{*}{\rotatebox[origin=c]{90}{RGB-D}}}    & Base-Oracle &  \bb{0.026}  &   0.398    &   0.336    &  0         &  0                \\
                                                                           & Base-SIFT   &   0.050      &   0.577    &   0.703    & 12.010     &  56.021           \\
                                                                           & Base-FF     &   0.045      &   0.548    &   0.613    &  4.709     &  46.058           \\
                                                                           & Base-Matlab &      -       &     -      &     -      &  12.831    &  49.612
                                                                           \\
                                                                           & Base-Mat-F  &     -        &     -      &     -      &   2.917    &  22.523
                                                                           \\
                                                                           & DeMoN       &   0.028      & \bb{0.130} & \bb{0.212} & \bb{2.641} &  \bb{20.585}      \\
      \hline                                                                                                                                                       
      \hline                                                                                                                                                                      
      \parbox[t]{2mm}{\multirow{6}{*}{\rotatebox[origin=c]{90}{Sun3D}}}    & Base-oracle &   0.020      &   0.241    &   0.220    & 0          &  0                \\
                                                                           & Base-SIFT   &   0.029      &   0.290    &   0.286    & 7.702      &  41.825           \\
                                                                           & Base-FF     &   0.029      &   0.284    &   0.297    & 3.681      &  33.301           \\
                                                                           & Base-Matlab &      -       &     -      &     -      & 5.920      &  32.298  
                                                                           \\
                                                                           & Base-Mat-F  &     -        &     -      &     -      &   2.230    &  26.338
                                                                           \\
                                                                           & DeMoN       &   \bb{0.019} & \bb{0.114} & \bb{0.172} & \bb{1.801} &  \bb{18.811}      \\
      \hline
      \hline                                                                                                                                                                      
      \parbox[t]{2mm}{\multirow{6}{*}{\rotatebox[origin=c]{90}{NYUv2}}}    & Base-oracle &     -        &     -      &     -      &     -      &    -              \\
                                                                           & Base-SIFT   &     -        &     -      &     -      &     -      &    -              \\
                                                                           & Base-FF     &     -        &     -      &     -      &     -      &    -              \\
                                                                           & Base-Matlab &     -        &     -      &     -      &     -      &    -             
                                                                           \\
                                                                           & Base-Mat-F  &     -        &     -      &     -      &     -      &    -
                                                                           \\
                                                                           & DeMoN       &     -        &     -      &     -      &     -      &    -              \\
      \hline

    \end{tabular}
    }
  \end{flushleft}
  \end{minipage}
  \begin{minipage}{0.15\linewidth}
  \begin{flushright}
    \vspace*{-.033cm}%
    \setlength{\tabcolsep}{0.1cm}
    { \scriptsize
      \begin{tabular}{|c|c|}
      \hline
                  & Depth        \\
      \hline
      Method      & sc-inv       \\
      \hline                     
      \hline                     
                  &              \\
                  &              \\
      Liu indoor  &  0.260       \\
      Liu outdoor &  0.341       \\
      Eigen VGG   &  0.225       \\
      DeMoN       & \bb{0.203}   \\
      \hline                     
      \hline   
                  &              \\
                  &              \\
      Liu indoor  &  0.816       \\
      Liu outdoor &  0.814       \\
      Eigen VGG   &  0.763       \\
      DeMoN       & \bb{0.303}   \\
      \hline                     
      \hline  
                  &              \\
                  &              \\
      Liu indoor  &  0.338       \\
      Liu outdoor &  0.428       \\
      Eigen VGG   &  0.272       \\
      DeMoN       & \bb{0.134}   \\
      \hline                     
      \hline  
                  &              \\
                  &              \\
      Liu indoor  &  0.214       \\
      Liu outdoor &  0.401       \\
      Eigen VGG   &  0.175       \\
      DeMoN       & \bb{0.126}   \\
      \hline
      \hline 
                  &              \\
                  &              \\
      Liu indoor  &  0.210       \\
      Liu outdoor &  0.421       \\
      Eigen VGG   & \bb{0.148}   \\
      DeMoN       &  0.180       \\
      \hline
    \end{tabular}
    }
  \end{flushright}
  \end{minipage}
  \vspace{0.3cm}
  \caption{
\bb{Left:} Comparison of two-frame depth and motion estimation methods. 
Lower is better for all measures. 
For a fair comparison with the baseline methods, we evaluate depth only at pixels visible in both images. 
For Base-Matlab depth is only available as a sparse point cloud and is therefore not compared to here.
We do not report the errors on NYUv2 since motion ground truth (and therefore depth scale) is not available. 
\bb{Right:} Comparison to single-frame depth estimation. Since the scale estimates are not comparable, we report only the scale invariant error metric.
}
  \label{tbl:results_unscaled}
\vspace{\figvspace}
\end{table}

\tab{tbl:results_unscaled} shows that DeMoN outperforms all baseline methods both on motion and depth accuracy by a factor of $1.5$ to $2$ on most datasets. 
The only exception is the MVS dataset where the motion accuracy of DeMoN is on par with the strong baseline based on FlowFields optical flow.
This demonstrates that traditional methods work well on the texture rich scenes present in MVS, but do not perform well for example on indoor scenes, with large homogeneous regions or small baselines where priors may be very useful.
Besides that, all Base-* methods use images at the full resolution of $640\times480$, while our method uses downsampled images of $256\times192$ as input.
\begin{NEW}
Higher resolution gives the Base-* methods an advantage in depth accuracy, but on the other hand these methods are more prone to outliers.
For detailed error distributions see the supplemental material.
\end{NEW}
Remarkably, on all datasets except for MVS the depth estimates of DeMoN are better than the ones a traditional approach can produce given the ground truth motion.
This is supported by qualitative results in \fig{fig:depth_comparison}.
We also note that DeMoN has smaller motion errors than (``Base-Mat-F''), showing its advantage over classical methods in motion estimation.

\begin{figure}
\begin{center}
  \begin{tikzpicture}[scale=.25]
    \node at (0,0) {\includegraphics[width=0.11\textwidth]{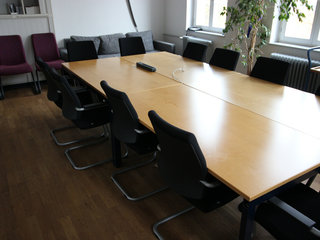}};
    \node at (8,0) {\includegraphics[width=0.11\textwidth]{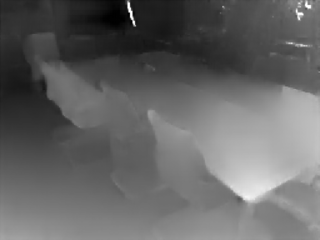}};
    \node at (16,0) {\includegraphics[width=0.11\textwidth]{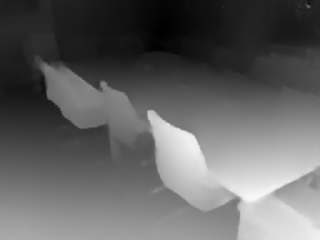}};
    \node at (24,0) {\includegraphics[width=0.11\textwidth]{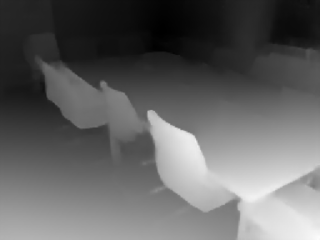}};
    \node at (8,-6.1) {\includegraphics[width=0.11\textwidth]{png/small_baseline/IMG_4352.png}};
    \node at (16,-6.1) {\includegraphics[width=0.11\textwidth]{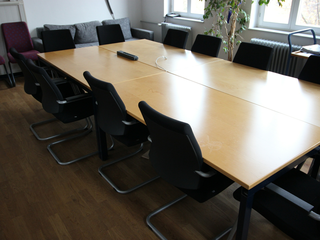}};
    \node at (24,-6.1) {\includegraphics[width=0.11\textwidth]{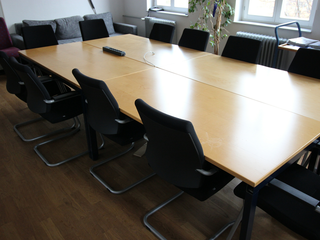}};
    \draw (-2.5,-7.5) -- ++(6,4);
    \node at (-0.9,-3.8) {\small Reference};
    \node at (3.1,-6.6) [rotate=90] {\small Second};
  \end{tikzpicture}
\end{center}
\vspace{\capvspace}%
\caption{%
Qualitative performance gain by increasing the baseline between the two input images for DeMoN. The depth map is produced with the top left reference image and the second image below. The first output is obtained with two identical images as input, which is a degenerate case for traditional structure from motion. 
}
\label{fig:small_baseline}
\vspace{\figvspace}%
\end{figure}

\begin{figure}
  \begin{center}
    \includegraphics[width=\linewidth]{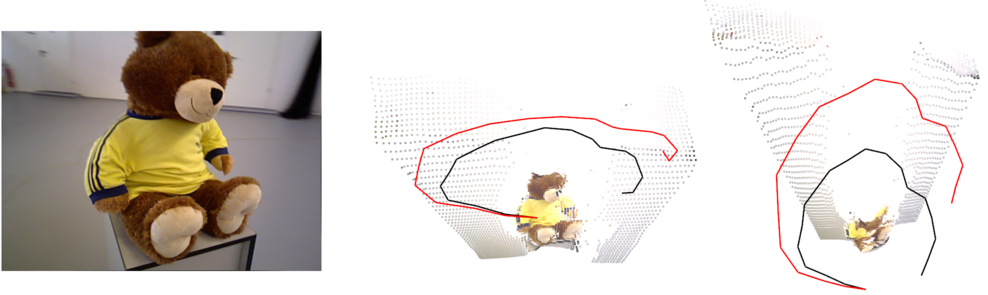}\\
    \vspace{-0.5em}
    \small First frame \hspace*{1.3cm} Frontal view \hspace*{1.1cm} Top view
  \end{center}
\vspace{\capvspace}%
\caption{%
Result on a sequence of the RGB-D SLAM dataset~\cite{sturm12iros}. 
The accumulated pairwise pose estimates by our network (red) are locally consistent with the ground truth trajectory (black).
The depth prediction of the first frame is shown. 
The network also separates foreground and background in its depth output.}
  \label{fig:trajectory_teddy}
\vspace{\figvspace}
\vspace{-0.3em}
\end{figure}

In contrast to classical approaches, we can also handle cases without and with very little camera motion, see Fig.~\ref{fig:small_baseline}.
  We used our network to compute camera trajectories by simple concatenation of the motion of consecutive frames, as shown in \fig{fig:trajectory_teddy}.
%We demonstrate how our approach can be used for computing camera trajectories in \fig{fig:trajectory_teddy} by simple concatenation of the motion of consecutive frames.
The trajectory shows mainly translational drift. 
We also did not apply any drift correction which is a crucial component in SLAM systems, but results convince us that DeMoN can be integrated into such systems.

\subsection{Comparison to depth from single image}

To demonstrate the value of the motion parallax, we additionally compare to the single-image depth estimation methods by Eigen \& Fergus~\cite{eigen_predicting_2015} and Liu et al.~\cite{liu_learning_2015}.
We compare to the improved version of the Eigen \& Fergus method, which is based on the VGG network architecture, and to two models by Liu et al.: one trained on indoor scenes from the NYUv2 dataset (``indoor'') and another, trained on outdoor images from the Make3D dataset~\cite{Saxena05learningdepth} (``outdoor'').

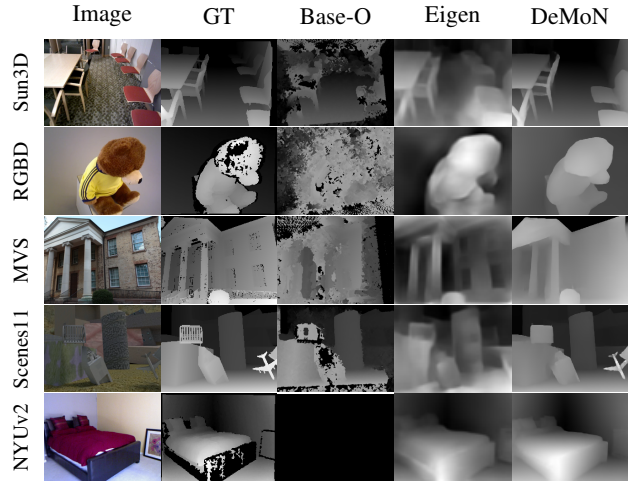
\begin{figure}
\begin{center}
\begin{tikzpicture}[xscale=1.55,yscale=1.18]
  %%Image,GT,Base-O,Eigen,DeMoN
  %%Sun3D,RGBD,MVS,Scenes11,NYUv2
  \node at (0,4.75) {\small Image};
  \node at (1,4.75) {\small GT};
  \node at (2,4.75) {\small Base-O};
  \node at (3.005,4.75) {\small Eigen};
  \node at (4.01,4.75) {\small DeMoN};

  \node at (-.7,4) [rotate=90] {\footnotesize Sun3D};
  \node at (-.7,3) [rotate=90] {\footnotesize RGBD};
  \node at (-.7,2) [rotate=90] {\footnotesize MVS};
  \node at (-.7,1) [rotate=90] {\footnotesize Scenes11};
  \node at (-.7,0) [rotate=90] {\footnotesize NYUv2};

  \node at (0,4) {\includegraphics[width=0.088\textwidth]{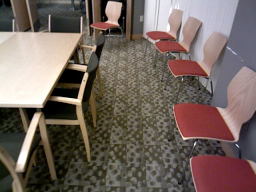}};
  \node at (1,4) {\includegraphics[width=0.088\textwidth]{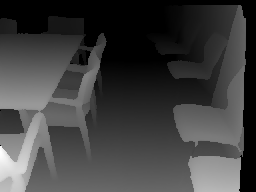}};
  \node at (2,4) {\includegraphics[width=0.088\textwidth]{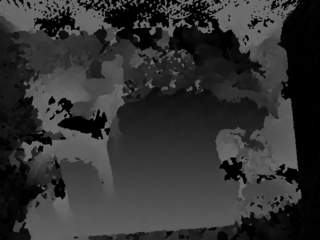}};
  \node at (3.005,4) {\includegraphics[width=0.089\textwidth]{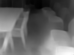}};
  \node at (4.01,4) {\includegraphics[width=0.088\textwidth]{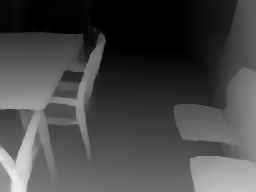}};
85
  \node at (0,3) {\includegraphics[width=0.088\textwidth]{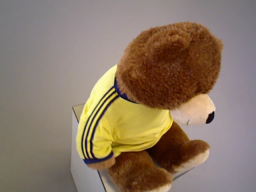}};
  \node at (1,3) {\includegraphics[width=0.088\textwidth]{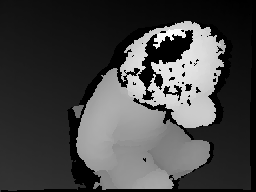}};
  \node at (2,3) {\includegraphics[width=0.088\textwidth]{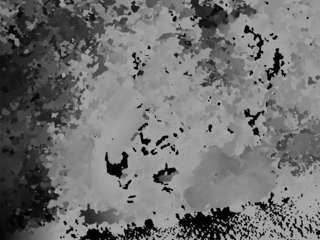}};
  \node at (3.005,3) {\includegraphics[width=0.089\textwidth]{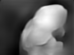}};
  \node at (4.01,3) {\includegraphics[width=0.088\textwidth]{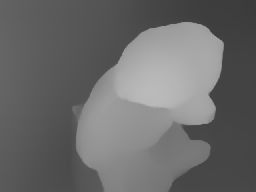}};
85
  \node at (0,2) {\includegraphics[width=0.088\textwidth]{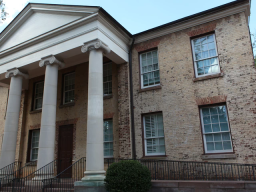}};
  \node at (1,2) {\includegraphics[width=0.088\textwidth]{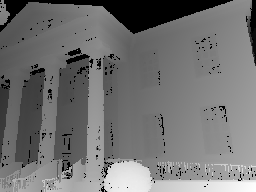}};
  \node at (2,2) {\includegraphics[width=0.088\textwidth]{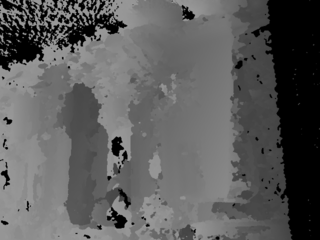}};
  \node at (3.005,2) {\includegraphics[width=0.089\textwidth]{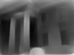}};
  \node at (4.01,2) {\includegraphics[width=0.088\textwidth]{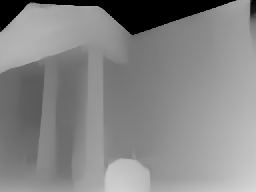}};
85
  \node at (0,1) {\includegraphics[width=0.088\textwidth]{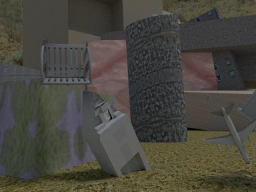}};
  \node at (1,1) {\includegraphics[width=0.088\textwidth]{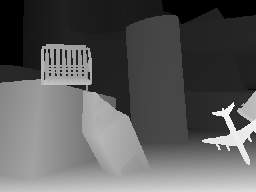}};
  \node at (2,1) {\includegraphics[width=0.088\textwidth]{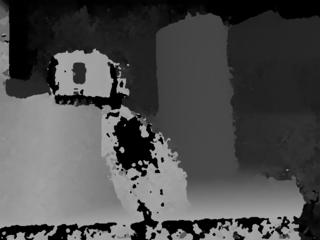}};
  \node at (3.005,1) {\includegraphics[width=0.089\textwidth]{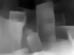}};
  \node at (4.01,1) {\includegraphics[width=0.088\textwidth]{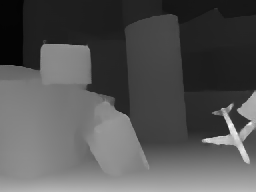}};
85
  \node at (0,0) {\includegraphics[width=0.088\textwidth]{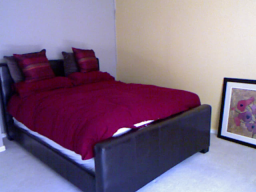}};
  \node at (1,0) {\includegraphics[width=0.088\textwidth]{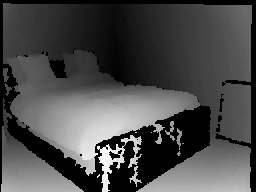}};
  \node at (2,0) {\includegraphics[width=0.088\textwidth]{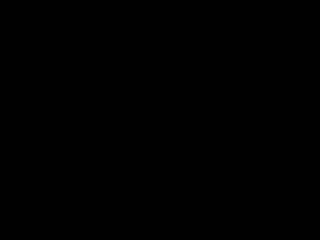}};
  \node at (3.005,0) {\includegraphics[width=0.089\textwidth]{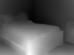}};
  \node at (4.01,0) {\includegraphics[width=0.088\textwidth]{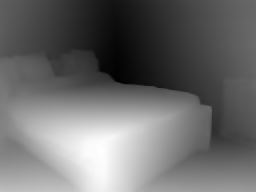}};
\end{tikzpicture}
\end{center}
\vspace{-0.4em}%
\vspace{\capvspace}%
\caption{
\bb{Top:} Qualitative depth prediction comparison on various datasets. The predictions of DeMoN are very sharp and detailed.
The Base-Oracle prediction on NYUv2 is missing because the motion ground truth is not available. 
Results on more methods and examples are shown in the supplementary material. %\bb{Bottom:} Visualization of the points clouds on NYUv2 example.
% For Liu's method the indoor model is used for Sun3d, RGBD and NYUv2 while for MVS and synthetic datasets the results of outdoor model are selected. 
% In the MVS example single image methods present an obvious wrong depth prediction on the windows of the building and in Scenes11 Liu's method tends to fit some indoor objects to the sythetic scenes. It can be seen that our method yields a clearly better result on both indoor and outdoor datasets. 
%\tcj{The names of the datasets are very hard to read..}
}
\label{fig:depth_comparison}
\vspace{\figvspace}
\vspace{-0.5em}
\end{figure}

The comparison in \fig{fig:depth_comparison} shows that the depth maps produced by DeMoN are more detailed and more regular than the ones produced by other methods. This becomes even more obvious when the results are visualized as a point cloud; see the videos in the supplemental material. 
%Even on NYUv2, on which the method of Eigen \& Fergus was trained, their point cloud is very noisy.

On all but one dataset, DeMoN outperforms the single-frame methods also by numbers, typically by a large margin. Notably, a large improvement can be observed even on the indoor datasets, Sun3D and RGB-D, showing that the additional stereopsis complements the other cues that can be learned from the large amounts of training data available for this scenario.
Only on the NYUv2 dataset, DeMoN is slightly behind the method of Eigen \& Fergus. 
This is because the comparison is not totally fair: the network of Eigen \& Fergus as well as Liu indoor was trained on the training set of NYUv2, whereas the other networks have not seen this kind of data before.

\vspace{-0.3em}%
\subsubsection{Generalization to new data}
\label{sec:generalization}
%\zlabel{zsec:generalization} % hyperreferenzing to external document doesn't work?

% \tca{
% The proposed method is intended to be applicable to image pairs coming from arbitrary types of scenes.
% This is in contrast with existing deep learning approaches to depth estimation which heavily rely on priors learned from scenes of specific types.
% Here we perform a targeted analysis of generalization capabilities of different approaches to previously unseen data.
% To this end we have compiled a small dataset of images showing uncommon or complicated scenes, for example abstract sculptures, close-ups of people and objects, images rotated by 90 degrees (more details are provided in the supplementary material). None of the compared approaches have been trained on images of this kind. \tab{tbl:generalization} reports the quantitative results of different methods on this dataset, while \fig{fig:generalization} shows a qualitative comparison.
% DeMoN significantly outperforms the competing approaches both quantitatively and qualitatively.}

Scene-specific priors learned during training may be useless or even harmful when being confronted with a scene that is very different from the training data. In contrast, the geometric relations between a pair of images are independent of the content of the scene and should generalize to unknown scenes. 
To analyze the generalization properties of DeMoN, we compiled a small dataset of images showing uncommon or complicated scenes, for example abstract sculptures, close-ups of people and objects, images rotated by 90 degrees.% More details are provided in the supplementary material.

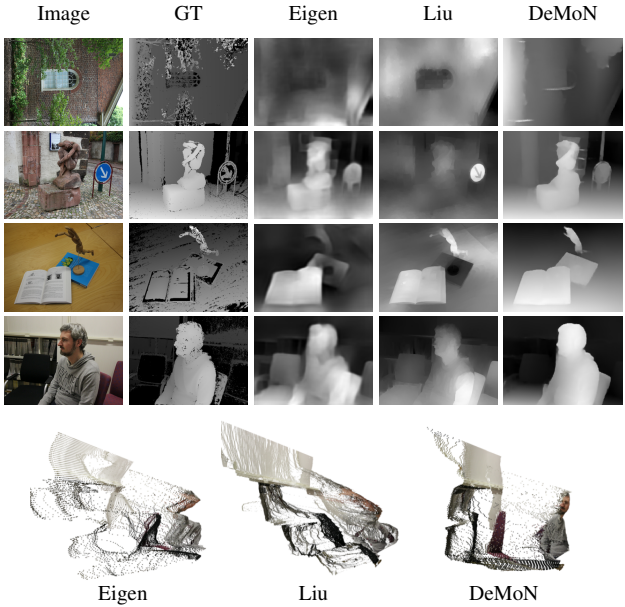
\begin{figure}
\begin{center}
\incgraphicsabove{width=0.09\textwidth,trim={0.04\Width} {0.045\Height} {0.04\Width} {0.045\Height},clip}{png/generalization/rgb_img/IMG_1285.png}{\footnotesize Image}
\incgraphicsabove{width=0.09\textwidth,trim={0.04\Width} {0.045\Height} {0.04\Width} {0.045\Height},clip}{png/generalization/gt_depth/IMG_1285_norm_img.png}{\footnotesize GT}
%\adjustimage{width=0.09\textwidth,trim={0.04\Width} {0.045\Height} {0.04\Width} {0.045\Height},clip]{png/generalization/ours/IMG_1285_IMG_1281_norm_img_2.png}
\incgraphicsabove{width=0.09\textwidth}{png/generalization/eigen/IMG_1285_norm_img_0_scaled.png}{\footnotesize Eigen}
\incgraphicsabove{width=0.09\textwidth,trim={0.04\Width} {0.045\Height} {0.04\Width} {0.045\Height},clip}{png/generalization/liu/IMG_1285_norm_img_0.png}{\footnotesize Liu}
\incgraphicsabove{width=0.09\textwidth,trim={0.04\Width} {0.045\Height} {0.04\Width} {0.045\Height},clip}{png/generalization/ours_refined/IMG_1285_IMG_1281_norm_img_2_ref.png}{\footnotesize DeMoN}\\
\vspace{0.1em}
\adjustimage{width=0.09\textwidth,trim={0.04\Width} {0.045\Height} {0.04\Width} {0.045\Height},clip}{png/generalization/rgb_img/IMG_4022.png}
\adjustimage{width=0.09\textwidth,trim={0.04\Width} {0.045\Height} {0.04\Width} {0.045\Height},clip}{png/generalization/gt_depth/IMG_4022_norm_img.png}
%\adjustimage{width=0.09\textwidth,trim={0.04\Width} {0.045\Height} {0.04\Width} {0.045\Height},clip}{png/generalization/ours/IMG_4022_IMG_4023_norm_img_2.png}
\adjustimage{width=0.09\textwidth}{png/generalization/eigen/IMG_4022_norm_img_0_scaled.png}
\adjustimage{width=0.09\textwidth,trim={0.04\Width} {0.045\Height} {0.04\Width} {0.045\Height},clip}{png/generalization/liu/IMG_4022_norm_img_0.png}
\adjustimage{width=0.09\textwidth,trim={0.04\Width} {0.045\Height} {0.04\Width} {0.045\Height},clip}{png/generalization/ours_refined/IMG_4022_IMG_4023_norm_img_2_ref.png}\\
\vspace{0.1em}
\adjustimage{width=0.09\textwidth,trim={0.04\Width} {0.045\Height} {0.04\Width} {0.045\Height},clip}{png/generalization/rgb_img/IMG_4058.png}
\adjustimage{width=0.09\textwidth,trim={0.04\Width} {0.045\Height} {0.04\Width} {0.045\Height},clip}{png/generalization/gt_depth/IMG_4058_norm_img.png}
%\adjustimage{width=0.09\textwidth,trim={0.04\Width} {0.045\Height} {0.04\Width} {0.045\Height},clip}{png/generalization/ours/IMG_4058_IMG_4059_norm_img_2.png}
\adjustimage{width=0.09\textwidth}{png/generalization/eigen/IMG_4058_norm_img_0_scaled.png}
\adjustimage{width=0.09\textwidth,trim={0.04\Width} {0.045\Height} {0.04\Width} {0.045\Height},clip}{png/generalization/liu/IMG_4058_norm_img_0.png}
\adjustimage{width=0.09\textwidth,trim={0.04\Width} {0.045\Height} {0.04\Width} {0.045\Height},clip}{png/generalization/ours_refined/IMG_4058_IMG_4059_norm_img_2_ref.png}\\
\vspace{0.1em}
\adjustimage{width=0.09\textwidth,trim={0.04\Width} {0.045\Height} {0.04\Width} {0.045\Height},clip}{png/generalization/rgb_img/IMG_4117.png}
\adjustimage{width=0.09\textwidth,trim={0.04\Width} {0.045\Height} {0.04\Width} {0.045\Height},clip}{png/generalization/gt_depth/IMG_4117_norm_img.png}
%\adjustimage{width=0.09\textwidth,trim={0.04\Width} {0.045\Height} {0.04\Width} {0.045\Height},clip}{png/generalization/ours/IMG_4117_IMG_4119_norm_img_2.png}
\adjustimage{width=0.09\textwidth}{png/generalization/eigen/IMG_4117_norm_img_0_scaled.png}
\adjustimage{width=0.09\textwidth,trim={0.04\Width} {0.045\Height} {0.04\Width} {0.045\Height},clip}{png/generalization/liu/IMG_4117_norm_img_0.png}
\adjustimage{width=0.09\textwidth,trim={0.04\Width} {0.045\Height} {0.04\Width} {0.045\Height},clip}{png/generalization/ours_refined/IMG_4117_IMG_4119_norm_img_2_ref.png}\\
\vspace{0.3em}
\newcommand{\incgraphicslabelinsidebottom}[3]{%
  \begin{tikzpicture}%
    \node[inner sep=0pt] (mynode) {\adjustimage{#1}{#2}};%
    \node at (mynode.south) [inner sep=0,scale=1,above=-3mm] {\strut #3};%
    %\draw [blue] (current bounding box.south west) rectangle (current bounding box.north east);%
  \end{tikzpicture}%
}
\incgraphicslabelinsidebottom{width=0.14\textwidth}{png/generalization/point_clouds/close_benjamin_eigen_white.png}{\footnotesize Eigen}
\incgraphicslabelinsidebottom{width=0.14\textwidth}{png/generalization/point_clouds/close_benjamin_liu_indoor_white.png}{\footnotesize Liu}
\incgraphicslabelinsidebottom{width=0.14\textwidth}{png/generalization/point_clouds/close_benjamin_ours_refined_white.png}{\footnotesize DeMoN}
\end{center}
\vspace{-0.4em}%
\vspace{\capvspace}%
\caption{%
Visualization of DeMoN's generalization capabilities to previously unseen configurations. Single-frame methods have severe problems in such cases, as most clearly visible in the point cloud visualization of the depth estimate for the last example.  %From left to right: RGB image, inverse
%depth of GT, DeMoN, Liu et al. \cite{liu_learning_2015}, and Eigen et al. %\cite{eigen_predicting_2015}.
%\TODO{Caption doesn't match anymore!}
}
\label{fig:generalization}
\vspace{\figvspace}
\vspace{0.2em}
\end{figure}

\fig{fig:generalization} and \tab{tbl:generalization} show that DeMoN, as to be expected, generalizes better to these unexpected scenes than single-image methods. It shows that the network has learned to make use of the motion parallax.

%\tch{can we find a way to merge figure 7 and 8?}

\begin{table}
   \begin{center}
     \setlength{\tabcolsep}{0.06cm}
     { \small
     \begin{tabular}{|c|ccc|}
       \hline
       Method             &   L1-inv    &   sc-inv     &   L1-rel     \\
       \hline
       Liu \cite{liu_learning_2015}     &    0.055    &    0.247       &    0.194 \\
       Eigen \cite{eigen_predicting_2015} &    0.062  &    0.238       &    0.185 \\
       DeMoN              & \bf{0.041}  & \bf{0.183}  & \bf{0.130}              \\
       \hline
     \end{tabular}
         % \begin{tabular}{|c|ccc|ccc|}
    %   \hline
    %   Method             &   L1-inv    &   sc-inv     &   L1-rel    &  {\tiny $\delta \!\!<\!\! 1.25$} &  {\tiny $\delta \!\!<\!\! 1.25^2$} &  {\tiny $\delta \!\!<\!\! 1.25^3$} \\
    %   \hline
    %   Liu \cite{liu_learning_2015}     &    0.055    &    0.247     &    0.194    &             0.713    &         0.911    &         0.957    \\
    %   Eigen \cite{eigen_predicting_2015} &    0.062    &    0.238     &    0.185  &       0.725    &         0.915    &         0.971    \\
    %   DeMoN              & \bf{0.041}  & \bf{0.183}   & \bf{0.130}    &    \bf{0.853}  &      \bf{0.950}  &      \bf{0.980}  \\
    %   \hline
    % \end{tabular}
    }
   \end{center}
\vspace{\capvspace}%
\vspace{0.2em}%
   \caption{%
   Quantitative generalization performance on previously unseen scenes, objects, and camera rotations, 
using a self-recorded and reconstructed dataset. 
Errors after optimal log-scaling. The best model of Eigen et al. \cite{eigen_predicting_2015} for this task is based on VGG, for Liu et al. \cite{liu_learning_2015}, the model trained on Make3D \cite{make3d_2009} performed best. DeMoN achieved best performance after two iterations.
   }%
   \label{tbl:generalization}
\vspace{\figvspace}
\vspace{-0.4em}
 \end{table}

% \begin{table}
%   \begin{center}
%     \setlength{\tabcolsep}{0.06cm}
%     { \small
    % \begin{tabular}{|c|ccc|ccc|}
    %   \hline
    %   Method             &   L1-inv    &   sc-inv     &   L1-rel    &  {\tiny $\delta \!\!<\!\! 1.25$} &  {\tiny $\delta \!\!<\!\! 1.25^2$} &  {\tiny $\delta \!\!<\!\! 1.25^3$} \\
    %   \hline
    %   Liu \cite{liu_learning_2015}     &    0.055    &    0.247     &    0.194    &             0.713    &         0.911    &         0.957    \\
    %   Eigen \cite{eigen_predicting_2015} &    0.062    &    0.238     &    0.185  &       0.725    &         0.915    &         0.971    \\
    %   DeMoN              & \bf{0.041}  & \bf{0.183}   & \bf{0.130}    &    \bf{0.853}  &      \bf{0.950}  &      \bf{0.980}  \\
    %   \hline

    % \end{tabular}
    % }
%   \end{center}
%   \caption{%
%   Quantitative generalization performance on previously unseen scenes, objects, and camera rotations, using a self-recorded and reconstructed dataset. Errors after optimal log-scaling. The best model of Eigen et al. \cite{eigen_predicting_2015} for this task is based on VGG, for Liu et al. \cite{liu_learning_2015}, the model trained on Make3D \cite{make3d_2009} performed best. DeMoN achieved best performance after two iterations.
  %\TODO{did we define the delta metrics anywhere?}
%   }
%   \label{tbl:generalization}
% \end{table}

\subsection{Ablation studies}
\label{sec:ablation_study}
Our architecture contains some design decisions that we justify by the following ablation studies. 
%First, we show that stacking two encoder-decoder networks with different tasks is crucial for estimating depth and camera motion from two images.
%Second, we show the influence of the scale invariant gradient loss function and additional predictions such as optical flow confidence and surface normals.
All results have been obtained on the Sun3D dataset with the bootstrap net.

%\paragraph{Na{\"i}ve approach vs stacking.}
%A single encoder-decoder network trained for depth prediction from two images degenerates to a solution using only the first image.
%As can be seen in Table \ref{tbl:ablation_naive_two_frame}, there is no significant difference between a network which predicts the depth from a single image and the two image version.
%It is well known that neural networks often tend to exploit simple solutions. 
%In our case, even when presented with two images, the network prefers to infer the depth from scene knowledge acquired during training, rather than from correspondences between the two images.
%By explicitly predicting optical flow between the two images we force the network to find correspondences, which results in a large performance improvement.
% compare with single depth AND motion here?

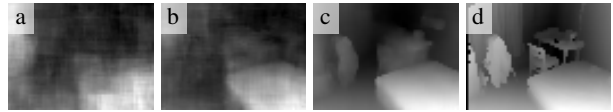
\begin{figure}
\newcommand{\incgraphicslabelinsidetopleft}[3]{%
  \begin{tikzpicture}%
    \node[inner sep=0pt] (mynode) {\adjustimage{#1}{#2}};%
    \node at (mynode.north west) [inner sep=0pt,minimum width=1em,scale=1,below right=-0.5pt,fill=white,fill opacity=0.6, text opacity=1] {\strut #3};%
    %\draw [blue] (current bounding box.south west) rectangle (current bounding box.north east);%
  \end{tikzpicture}%
}%
\begin{center}
  \incgraphicslabelinsidetopleft{width=0.11\textwidth}{png/grad/nogradient/0_predicted_depth.png}{\small a}
  \incgraphicslabelinsidetopleft{width=0.11\textwidth}{png/grad/normal/0_predicted_depth.png}{\small b}
  \incgraphicslabelinsidetopleft{width=0.11\textwidth}{png/grad/gradient/0_predicted_depth.png}{\small c}
  \incgraphicslabelinsidetopleft{width=0.11\textwidth}{png/grad/gradient/0_depth.png}{\small d}%
\end{center}
\vspace{\capvspace}%
\caption{%
Depth prediction comparison with different outputs and losses.
\bb{(a)} Just L1 loss on the absolute depth values.
\bb{(b)} Additional output of normals and L1 loss on the normals.
\bb{(c)} Like (b) but with the proposed gradient loss.
\bb{(d)} Ground truth.
}
\label{fig:gradient_comparison}
\vspace{\figvspace}
\end{figure}

\bb{Choice of the loss function.}
%\tcb{I'm a bit confused about the experiment with the optical flow. How can you train this network without a loss on optical flow?}
\tab{tbl:ablation_loss} shows the influence of the loss function on the accuracy of the estimated depth and motion. Interestingly, while the scale invariant loss greatly improves the prediction qualitatively (see \fig{fig:gradient_comparison}), it has negative effects on depth scale estimation. 
This leads to weak performance on non-scale-invariant metrics and the motion accuracy. 
%\tcb{But it also affects the camera motion estimation, although this should be scale-invariant?}
% \tab{tbl:ablation_loss} also shows that 
Estimation of the surface normals slightly improves all results.
Finally, the full architecture with the scale invariant loss, normal estimation, and a loss on the flow, leads to the best results. 
%\tca{do we have no no yes?}\tcb{no yes yes seems more important, since the gradient loss is bad?}

\bb{Flow confidence.}
Egomotion estimation only requires sparse but high-quality correspondences. \tab{tbl:ablation_confidence} shows that given the same flow, egomotion estimation improves when given the flow confidence as an extra input. Our interpretation is that the flow confidence helps finding most accurate correspondences. 
% \paragraph{Encoder-decoder pairs.}
% To force the network to relate both images we add an encoder-decoder network which predicts dense correspondences in the form of an optical flow field.
% We use the optical flow prediction as an additional input to our depth and motion encoder-decoder.

\begin{table}
  \begin{center}
    \setlength{\tabcolsep}{0.1cm}
    { \small
    \begin{tabular}{|ccc||ccc|cc|}
      \hline
            & &  & \multicolumn{3}{|c|}{Depth} & \multicolumn{2}{|c|}{Motion} \\
      \hline
      grad & norm & flow &  L1-inv &  sc-inv &  L1-rel &  rot &  tran  \\
      \hline
%       \no & \no   & \yes  &     - &     - &     - &        - &         - &    0.028 \\
      \no    & \no   & \no   &   0.040 &   0.211 &   0.354 &  3.127 &  30.861    \\
      \yes   & \no   & \no   &   0.057 &   0.159 &   0.437 &  4.585 &  39.819   \\
      \no    & \yes  & \no   &   0.037 &   0.190 &   0.336 &  2.570 &  29.607   \\
%       \yes   & \yes  & \no   &   0.058 &   0.163 &   0.603 &  4.472 &  41.766   \\
      \no    & \yes  & \yes  & \bb{0.029} &   0.184 &  \bb{0.266} &  \bb{2.359} &  \bb{23.578}   \\
      \yes   & \yes  & \yes  & 0.032 & \bb{0.150} & 0.276 & 2.479 & 24.372 \\
      \hline
    \end{tabular}
    }
  \end{center}
\vspace{\capvspace}%
\vspace{0.2em}%
  \caption{%
The influence of the loss function on the performance. 
The gradient loss improves the scale invariant error, but degrades the scale-sensitive measures. 
Surface normal prediction improves the depth accuracy. 
A combination of all components leads to the best tradeoff.
  }
  \label{tbl:ablation_loss}
\vspace{\figvspace}
\end{table}

\begin{table}
  \begin{center}
    \setlength{\tabcolsep}{0.1cm}
    { \small
    \begin{tabular}{|c||ccc|cc|c|}
      \hline
                 & \multicolumn{3}{c|}{Depth} & \multicolumn{2}{c|}{Motion} & Flow \\
      \hline
      Confidence &  L1-inv &  sc-inv &  L1-rel &  rot &  tran &  EPE  \\
      \hline
      \no    &  0.030 &  0.028 &  0.26 &     2.830 &      25.262 &    0.027 \\
      \yes   &  0.032 &  0.027 &  0.28 &     2.479 &      24.372 &    0.027 \\
    \hline
    \end{tabular}
    }
  \end{center}
\vspace{\capvspace}%
\vspace{0.2em}%
  \caption{The influence of confidence prediction on the overall performance of the different outputs.}
%   \vspace{-1ex}
  \label{tbl:ablation_confidence}
\vspace{\figvspace}
\vspace{-0.5em}
\end{table}

% \begin{table}
%   \begin{center}
%     \setlength{\tabcolsep}{0.1cm}
%     { \small
%     \begin{tabular}{|cc||ccc|cc|c|}
%       \hline
%       \multicolumn{2}{|c||}{Ground truth} & \multicolumn{3}{c|}{Depth} & \multicolumn{2}{c|}{Motion} & Flow \\
%       \hline
%       flow & dep+mot &  L1-inv &  sc-inv &  L1-rel &  rot    &  tran &  EPE \\
%       \hline
%       \yes & \no     &   0.007 &   0.058 &   0.066 &     -   &      -  &    -   \\
%       \yes & \no     &      -  &      -  &      -  &  0.340  &   2.235 &    -   \\
%       \no  & \no     &   0.058 &   0.163 &   0.603 &  4.472  &  41.766 &    -    \\ \hline
%       \no  & \yes    &      -  &      -  &    -    &    -    &   -     &  0.005 \\
%       \no  & \no     &   -     &  -      &   -     &    -    &  -      &  0.027 \\
%       \hline
%     \end{tabular}
%     }
%   \end{center}
%   \caption{\tcb{This table should probably go to the supplemental material. We also don't refer to it in the text, so far.} The effect of providing the ground truth optical flow (flow) or depth and motion (dep+mot) to the network. A network can be trained to produce very accurate depth and motion estimates given the ground truth optical flow, and vice-versa, a network can estimate the optical flow very well given the ground truth depth and motion.}
%   \label{tbl:ablation_groundtruth}
% \end{table}

\section{Conclusions and Future Work}
% graceful decay with increasing difficulty?
% single method that solves motion and depth
% problems/future work: small images, fixed camera intrinsics, ... probably a lot more
DeMoN is the first deep network that has learned to estimate depth and camera motion from two unconstrained images. 
Unlike networks that estimate depth from a single image, DeMoN can exploit the motion parallax, which is a powerful cue that generalizes to new types of scenes, and allows to estimate the egomotion.  
This network outperforms traditional structure from motion techniques on two frames, since in contrast to those, it is trained end-to-end and learns to integrate other shape from X cues. 
%Yet, it has not reached the flexibility with respect to the intrinsic parameters as traditional approaches.
When it comes to handling cameras with different intrinsic parameters it has not yet reached the flexibility of classic approaches.
The next challenge is to lift this restriction and extend this work to more than two images.
As in classical techniques, this is expected to significantly improve the robustness and accuracy.
%The general architecture of DeMoN is already well prepared for feeding more images in an iterative manner, and its ability to exploit the motion parallax between two images is the major requirement for a multi-frame network to be successful.  

\vspace{-0.7em}
\paragraph{Acknowledgements}
We acknowledge funding by the ERC Starting Grant VideoLearn, the DFG 
grant BR-3815/5-1, and the EU project Trimbot2020.

\clearpage

{\small
\bibliographystyle{ieee}
\bibliography{DepthMotionNet}

\begin{thebibliography}{10}\itemsep=-1pt

\bibitem{agarwal_ceres}
S.~Agarwal, K.~Mierle, and {Others}.
\newblock Ceres {Solver}.

\bibitem{agrawal_egomotion_2015}
P.~Agrawal, J.~Carreira, and J.~Malik.
\newblock Learning to see by moving.
\newblock In {\em IEEE {International} {Conference} on {Computer} {Vision}
  (ICCV)}, Dec. 2015.

\bibitem{bailer_flow_2015}
C.~Bailer, B.~Taetz, and D.~Stricker.
\newblock Flow {Fields}: {Dense} {Correspondence} {Fields} for {Highly}
  {Accurate} {Large} {Displacement} {Optical} {Flow} {Estimation}.
\newblock In {\em IEEE {International} {Conference} on {Computer} {Vision}
  (ICCV)}, Dec. 2015.

\bibitem{collins_space-sweep_1996}
R.~T. Collins.
\newblock A space-sweep approach to true multi-image matching.
\newblock In {\em Proceedings {{CVPR}} '96, 1996 {{IEEE Computer Society
  Conference}} on {{Computer Vision}} and {{Pattern Recognition}}, 1996}, pages
  358--363, June 1996.

\bibitem{dosovitskiy_flownet_2015}
A.~Dosovitskiy, P.~Fischer, E.~Ilg, P.~H{\"a}usser, C.~Haz{\i}rba{\c{s}},
  V.~Golkov, P.~v.d. Smagt, D.~Cremers, and T.~Brox.
\newblock Flownet: Learning optical flow with convolutional networks.
\newblock In {\em IEEE International Conference on Computer Vision (ICCV)},
  Dec. 2015.

\bibitem{Exemplar_CNN_PAMI}
A.~Dosovitskiy, P.~Fischer, J.~T. Springenberg, M.~Riedmiller, and T.~Brox.
\newblock Discriminative unsupervised feature learning with exemplar
  convolutional neural networks.
\newblock {\em IEEE Transactions on Pattern Analysis and Machine Intelligence},
  38(9):1734--1747, Oct 2016.
\newblock TPAMI-2015-05-0348.R1.

\bibitem{eigen_predicting_2015}
D.~Eigen and R.~Fergus.
\newblock Predicting {Depth}, {Surface} {Normals} and {Semantic} {Labels}
  {With} a {Common} {Multi}-{Scale} {Convolutional} {Architecture}.
\newblock In {\em {IEEE} {International} {Conference} on {Computer} {Vision}
  ({ICCV})}, Dec. 2015.

\bibitem{eigen_predicting_2014}
D.~Eigen, C.~Puhrsch, and R.~Fergus.
\newblock {Depth} {Map} {Prediction} from a {Single} {Image} using a
  {Multi}-{Scale} {Deep} {Network}.
\newblock In Z.~Ghahramani, M.~Welling, C.~Cortes, N.~D. Lawrence, and K.~Q.
  Weinberger, editors, {\em Advances in Neural Information Processing Systems
  27}, pages 2366--2374. Curran Associates, Inc., 2014.

\bibitem{engel14eccv}
J.~Engel, T.~Sch\"ops, and D.~Cremers.
\newblock {LSD-SLAM}: Large-scale direct monocular {SLAM}.
\newblock In {\em European Conference on Computer Vision (ECCV)}, September
  2014.

\bibitem{Faugeras1993}
O.~Faugeras.
\newblock {\em Three-dimensional Computer Vision: A Geometric Viewpoint}.
\newblock MIT Press, Cambridge, MA, USA, 1993.

\bibitem{fischler_random_1981}
M.~A. Fischler and R.~C. Bolles.
\newblock Random {Sample} {Consensus}: {A} {Paradigm} for {Model} {Fitting}
  with {Applications} to {Image} {Analysis} and {Automated} {Cartography}.
\newblock {\em Commun. ACM}, 24(6):381--395, June 1981.

\bibitem{flynn_deepstereo_2015}
J.~Flynn, I.~Neulander, J.~Philbin, and N.~Snavely.
\newblock Deepstereo: Learning to predict new views from the world's imagery.
\newblock In {\em Conference on Computer Vision and Pattern Recognition
  (CVPR)}, 2016.

\bibitem{make3d_2009}
D.~A. Forsyth.
\newblock Make3d: Learning 3d scene structure from a single still image.
\newblock {\em IEEE Transactions on Pattern Analysis and Machine Intelligence},
  31(5):824--840, May 2009.

\bibitem{frahm_building_2010}
J.-M. Frahm, P.~Fite-Georgel, D.~Gallup, T.~Johnson, R.~Raguram, C.~Wu, Y.-H.
  Jen, E.~Dunn, B.~Clipp, S.~Lazebnik, and M.~Pollefeys.
\newblock Building {Rome} on a {Cloudless} {Day}.
\newblock In K.~Daniilidis, P.~Maragos, and N.~Paragios, editors, {\em European
  Conference on Computer Vision (ECCV)}, number 6314 in Lecture {Notes} in
  {Computer} {Science}, pages 368--381. Springer Berlin Heidelberg, 2010.

\bibitem{fuhrmann2014mve}
S.~Fuhrmann, F.~Langguth, and M.~Goesele.
\newblock Mve-a multiview reconstruction environment.
\newblock In {\em Proceedings of the Eurographics Workshop on Graphics and
  Cultural Heritage (GCH)}, volume~6, page~8, 2014.

\bibitem{hartley_defense_1997}
R.~I. Hartley.
\newblock In defense of the eight-point algorithm.
\newblock {\em IEEE Transactions on Pattern Analysis and Machine Intelligence},
  19(6):580--593, June 1997.

\bibitem{Hartley2004}
R.~I. Hartley and A.~Zisserman.
\newblock {\em Multiple View Geometry in Computer Vision}.
\newblock Cambridge University Press, ISBN: 0521540518, second edition, 2004.

\bibitem{hirschmuller_accurate_2005}
H.~Hirschm{\"u}ller.
\newblock Accurate and efficient stereo processing by semi-global matching and
  mutual information.
\newblock In {\em IEEE International Conference on Computer Vision and Pattern
  Recognition (CVPR)}, volume~2, pages 807--814, June 2005.

\bibitem{Jayaraman2015egomotion}
D.~Jayaraman and K.~Grauman.
\newblock {Learning image representations tied to egomotion}.
\newblock In {\em ICCV}, 2015.

\bibitem{jia_caffe:_2014}
Y.~Jia, E.~Shelhamer, J.~Donahue, S.~Karayev, J.~Long, R.~Girshick,
  S.~Guadarrama, and T.~Darrell.
\newblock Caffe: {Convolutional} {Architecture} for {Fast} {Feature}
  {Embedding}.
\newblock {\em arXiv preprint arXiv:1408.5093}, 2014.

\bibitem{kendall_modelling_2015}
A.~Kendall and R.~Cipolla.
\newblock Modelling {Uncertainty} in {Deep} {Learning} for {Camera}
  {Relocalization}.
\newblock In {\em International Converence on Robotics and Automation (ICRA)},
  2016.

\bibitem{kingma_adam:_2014}
D.~Kingma and J.~Ba.
\newblock Adam: {A} {Method} for {Stochastic} {Optimization}.
\newblock {\em arXiv:1412.6980 [cs]}, Dec. 2014.
\newblock arXiv: 1412.6980.

\bibitem{li_iterative_2016}
K.~Li, B.~Hariharan, and J.~Malik.
\newblock Iterative {Instance} {Segmentation}.
\newblock In {\em 2016 {IEEE} {Conference} on {Computer} {Vision} and {Pattern}
  {Recognition} ({CVPR})}, pages 3659--3667, June 2016.

\bibitem{liu_learning_2015}
F.~Liu, C.~Shen, G.~Lin, and I.~Reid.
\newblock Learning {Depth} from {Single} {Monocular} {Images} {Using} {Deep}
  {Convolutional} {Neural} {Fields}.
\newblock In {\em IEEE Transactions on Pattern Analysis and Machine
  Intelligence}, 2015.

\bibitem{longuet-higgins_computer_1981}
H.~C. Longuet-Higgins.
\newblock A computer algorithm for reconstructing a scene from two projections.
\newblock {\em Nature}, 293(5828):133--135, Sept. 1981.

\bibitem{lowe_distinctive_2004}
D.~G. Lowe.
\newblock Distinctive {Image} {Features} from {Scale}-{Invariant} {Keypoints}.
\newblock {\em International Journal of Computer Vision}, 60(2):91--110, Nov.
  2004.

\bibitem{lucas_iterative_1981}
B.~D. Lucas and T.~Kanade.
\newblock An {Iterative} {Image} {Registration} {Technique} with an
  {Application} to {Stereo} {Vision}.
\newblock In {\em Proceedings of the 7th {International} {Joint} {Conference}
  on {Artificial} {Intelligence} - {Volume} 2}, {IJCAI}'81, pages 674--679, San
  Francisco, CA, USA, 1981. Morgan Kaufmann Publishers Inc.

\bibitem{mayer_sceneflownet_2016}
N.~Mayer, E.~Ilg, P.~H{\"a}usser, P.~Fischer, D.~Cremers, A.~Dosovitskiy, and
  T.~Brox.
\newblock A large dataset to train convolutional networks for disparity,
  optical flow, and scene flow estimation.
\newblock In {\em IEEE International Conference on Computer Vision and Pattern
  Recognition (CVPR)}, 2016.

\bibitem{NYU}
P.~K. Nathan~Silberman, Derek~Hoiem and R.~Fergus.
\newblock Indoor {Segmentation} and {Support} {Inference} from {RGBD} {Images}.
\newblock In {\em European Conference on Computer Vision (ECCV)}, 2012.

\bibitem{newcombe_dtam:_2011}
R.~A. Newcombe, S.~Lovegrove, and A.~Davison.
\newblock {{DTAM}}: {{Dense}} tracking and mapping in real-time.
\newblock In {\em 2011 {{IEEE International Conference}} on {{Computer Vision}}
  ({{ICCV}})}, pages 2320--2327, 2011.

\bibitem{nister_efficient_2004}
D.~Nister.
\newblock An efficient solution to the five-point relative pose problem.
\newblock {\em IEEE Transactions on Pattern Analysis and Machine Intelligence},
  26(6):756--770, June 2004.

\bibitem{Saxena05learningdepth}
A.~Saxena, S.~H. Chung, and A.~Y. Ng.
\newblock Learning depth from single monocular images.
\newblock In {\em In NIPS 18}. MIT Press, 2005.

\bibitem{schoenberger2016sfm}
J.~L. Sch\"{o}nberger and J.-M. Frahm.
\newblock Structure-from-motion revisited.
\newblock In {\em IEEE Conference on Computer Vision and Pattern Recognition
  (CVPR)}, 2016.

\bibitem{schoenberger2016mvs}
J.~L. Sch\"{o}nberger, E.~Zheng, M.~Pollefeys, and J.-M. Frahm.
\newblock Pixelwise view selection for unstructured multi-view stereo.
\newblock In {\em European Conference on Computer Vision (ECCV)}, 2016.

\bibitem{Shi_1994_3266}
J.~Shi and C.~Tomasi.
\newblock Good features to track.
\newblock In {\em 1994 IEEE Conference on Computer Vision and Pattern
  Recognition (CVPR'94)}, pages 593 -- 600, 1994.

\bibitem{sturm12iros}
J.~Sturm, N.~Engelhard, F.~Endres, W.~Burgard, and D.~Cremers.
\newblock A benchmark for the evaluation of rgb-d slam systems.
\newblock In {\em Proc. of the International Conference on Intelligent Robot
  Systems (IROS)}, Oct. 2012.

\bibitem{szegedy_rethinking_2015}
C.~Szegedy, V.~Vanhoucke, S.~Ioffe, J.~Shlens, and Z.~Wojna.
\newblock Rethinking the {{Inception Architecture}} for {{Computer Vision}}.
\newblock {\em arXiv:1512.00567 [cs]}, Dec. 2015.

\bibitem{Tomasi91detectionand}
C.~Tomasi and T.~Kanade.
\newblock Detection and tracking of point features.
\newblock Technical report, International Journal of Computer Vision, 1991.

\bibitem{triggs_bundle_2000}
B.~Triggs, P.~McLauchlan, R.~Hartley, and A.~Fitzgibbon.
\newblock Bundle {{Adjustment}} — {{A Modern Synthesis Vision Algorithms}}:
  {{Theory}} and {{Practice}}.
\newblock In B.~Triggs, A.~Zisserman, and R.~Szeliski, editors, {\em Vision
  {{Algorithms}}: {{Theory}} and {{Practice}}}, volume 1883, pages 153--177.
  {Springer Berlin / Heidelberg}, Apr. 2000.

\bibitem{UB15}
B.~Ummenhofer and T.~Brox.
\newblock Global, dense multiscale reconstruction for a billion points.
\newblock In {\em IEEE International Conference on Computer Vision (ICCV)}, Dec
  2015.

\bibitem{valgaerts_dense_2012}
L.~Valgaerts, A.~Bruhn, M.~Mainberger, and J.~Weickert.
\newblock Dense versus {{Sparse Approaches}} for {{Estimating}} the
  {{Fundamental Matrix}}.
\newblock {\em International Journal of Computer Vision}, 96(2):212--234, Jan.
  2012.

\bibitem{wu_towards_2013}
C.~Wu.
\newblock Towards {Linear}-{Time} {Incremental} {Structure} from {Motion}.
\newblock In {\em {International} {Conference} on 3D {Vision} (3DV)}, pages
  127--134, June 2013.

\bibitem{xiao_sun3d_2013}
J.~Xiao, A.~Owens, and A.~Torralba.
\newblock {SUN3D}: {A} {Database} of {Big} {Spaces} {Reconstructed} {Using}
  {SfM} and {Object} {Labels}.
\newblock In {\em IEEE International Conference on Computer Vision (ICCV)},
  pages 1625--1632, Dec. 2013.

\bibitem{zagoruyko_2015}
S.~Zagoruyko and N.~Komodakis.
\newblock Learning to compare image patches via convolutional neural networks.
\newblock In {\em Conference on Computer Vision and Pattern Recognition
  (CVPR)}, 2015.

\bibitem{zbontar_computing_2015}
J.~{\v{Z}}bontar and Y.~LeCun.
\newblock Computing the {Stereo} {Matching} {Cost} {With} a {Convolutional}
  {Neural} {Network}.
\newblock In {\em {IEEE} {Conference} on {Computer} {Vision} and {Pattern}
  {Recognition} ({CVPR})}, June 2015.

\end{thebibliography}
}

\IfFileExists{./depthmotionnet_supplement.pdf}{%
\clearpage


\section*{Supplementary material}

# DeMoN: Depth and Motion Network for Learning Monocular Stereo
## – Supplementary Material –

## A. Network Architecture Details

Our network is a chain of encoder-decoder networks. Figures 15 and 16 explain the details of the two encoder-decoders used in the bootstrap and iterative net part. Fig. 17 gives implementation details for the refinement net.

The encoder-decoders for the bootstrap and iterative net use additional inputs which come from previous predictions. Some of these inputs, like warped images or depth from optical flow, need to be generated with special layers, which we describe here.

**Warped second image** We warp the second image using the predicted or generated optical flow field. We compute all values with bilinear interpolation and fill values that fall outside the image boundaries with zeros.

After warping with ground truth optical flow, the second image should look like the first image with the exception of occlusion regions. Comparing first image and warped image allows to assess the quality of the optical flow.

**Flow from depth and motion** We generate an optical flow estimate based on a depth and camera motion estimate for the first encoder-decoder of the iterative net. This optical flow can be used as an initial estimate, which can be further improved by the iterative net.

**Depth from optical flow and motion** In contrast to generating optical flow from a depth map and camera motion, computing depth from optical flow and camera motion is not straightforward. To compute the depth value of a pixel, we first project the corresponding point in the second image to the epipolar line and then triangulate the 3D point to retrieve the depth.

This generated depth map provides an estimate which combines optical flow and camera motion. Note that using the estimated camera motion here ensures that the depth values of the estimate are correctly scaled according to the camera motion. In case of ground truth optical flow and camera motion this yields the true depth map.

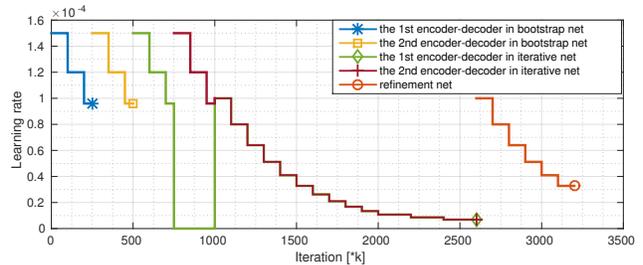

Figure 1. Learning rate schedule. During the training of each encoder-decoder a multistep learning rate is applied. At each step the learning rate is reduced by 20%.

## B. Training Schedule

We train our model from scratch for 3200k iterations in total. Fig. 1 shows our learning rate schedule. From 0 to 1000k iterations we sequentially train the four encoder-decoders, then the two encoder-decoders of the iterative net are jointly trained to iteration 2600k, finally we train the refinement net for another 600k iterations. This training schedule can be further optimized.

We normalize all point-wise losses with the number of pixels. The loss weights for flow, flow confidence, depth and normals are 1000, 1000, 300 and 100 respectively. The loss weight for the scale invariant gradient loss on flow is 1000. For depth we weight the scale invariant gradient loss with 1500, because we consider sharp edges and smooth surfaces more important for depth prediction. The translation and rotation losses are weighted with a factor of 15 and 160 respectively to balance their importance with respect to the other losses.

## C. Datasets

Our training procedure requires ground truth camera poses and ground truth depth for the first image.

The datasets we use for training can be divided in two groups: synthetic and real datasets. Real datasets provide natural images, but do only provide sparse pseudo ground truth due to measurement and reconstruction errors. Synthetic datasets provide perfect ground truth but often fea-

1

ture unrealistic images. Since no single dataset is perfect, we train on a set of five datasets with different properties. Tab. 1 provides an overview of the properties of the datasets. Figures 11, 12, 13, 14 and 3 show examples from the datasets we used.

**SUN3D** The dataset from [11] is a set of video sequences with camera poses and depth maps. Camera poses and depth maps are inaccurate because they have been created via a reconstruction process. The raw sensor data for this dataset was recorded with a structured light depth sensor which reports absolute depth values, thus the reconstruction scale for this dataset is known. The dataset features indoor scenes of offices, apartments, hotel rooms and university buildings.

To sample image pairs suitable for training, we apply a simple filtering process. To avoid image pairs with large pose or depth error, we filter out pairs with a large photo-consistency error. We also filter out image pairs with less than 50% of the pixels from the first image visible in the second image. Further we discard images with a baseline smaller than 5cm.

We have split the datasets into training and testing scenes. The training set contains 253 scenes and the test set contains 16 scenes. For training we sample a total of 117768 image pairs from all image sequences. For testing we generate a diverse set of 80 image pairs which minimizes overlap within image sequences. This way we obtain a small test set with large diversity.

**RGB-D SLAM** The dataset from [9] is a set of video sequences with camera poses and depth maps. Camera poses are accurate since they come from an external motion tracking system. The scale of the reconstructed camera poses is known. Depth maps have been recorded with a structured light sensor and are therefore affected by measurement noise and limited to a fixed depth range. Due to the use of an external tracking system, the data is restricted to indoor scenes in a lab environment.

We use the same sampling process as for SUN3D. The training set contains 45276 image pairs, while the test set features 80 diverse image pairs.

**MVS** In contrast to SUN3D and RGB-D SLAM, MVS is a dataset with outdoor images. We use the Citywall and Achteckturm datasets provided with [4] and the Breisach dataset from [10] for training. We compute depth maps and camera poses with the Multiview Reconstruction Environment by [4].

For testing we use datasets provided with the COLMAP software from Schönberger *et al*. [7, 8], which are called Person-Hall, Graham-Hall, and South-Building. These

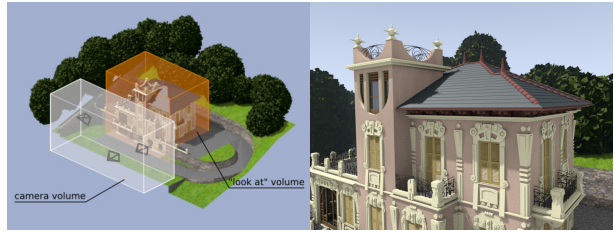

Figure 2. Annotation volumes for the Blendswap dataset. **Left:** We annotate each scene with volumes that describe valid camera positions (white box) and valid positions for the "look at" target (orange box). **Right:** Image from a randomly sampled camera and target.

datasets show the exterior of the eponymous buildings. Further, we have recorded a small scene which shows a sculpture and can be seen in the third row of Fig. 13. We use the COLMAP software for computing the depth and camera poses for these scenes.

Due to the scale ambiguity in the reconstruction process, the absolute scale is unknown. The datasets are small and not very diverse. Again we use the same image pair sampling strategy as for the previous datasets. We sample 16152 image pairs for training and 66 image pairs for testing.

**Scenes11** Scenes11 is a synthetic dataset with randomly generated geometry and objects from ShapeNet [3]. All scenes have a ground plane which makes the scenes look like outdoor scenes. We use the open source software Blender [2] for rendering.

Due to the synthetic nature of the dataset, the accuracy of the camera poses and depth maps is perfect. Disadvantages are the artificial look and a simplistic model of the camera movement. We sample camera parameters from Gaussian and uniform distributions, which bears the risk of learning the underlying model by heart when training only on this data.

The absolute scale for this dataset is meaningless since the scale of the objects is arbitrary and inconsistent. We generate 239508 image pairs from 19959 unique scenes for training and 128 image pairs from 128 unique scenes for testing.

**Blendswap** The blendswap dataset is an artificial dataset generated with 150 scenes from `blendswap.com`. Scenes range from cartoon-like to photorealistic. Blendswap contains some outdoor scenes, but a large part of the scenes shows interiors. We annotated each of the 150 scenes with a camera volume, which marks free space that can be used for camera placement, and a "look at" volume for placing target points defining the

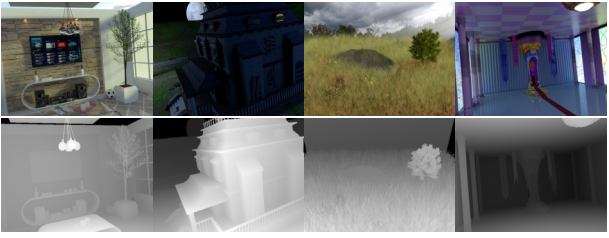

Figure 3. Images and the corresponding depth maps from the Blendswap dataset. Blendswap contains 150 distinct scenes, with a large variety of different styles and settings.

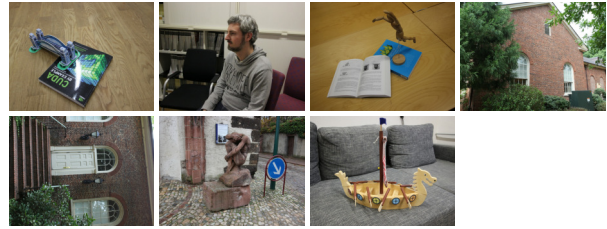

Figure 4. Samples of the seven scenes used for the generalization experiment.

camera viewing direction. We sample camera positions and targets uniformly over the annotated volumes. Fig. 2 shows an example of an annotated scene and an automatically generated image. When sampling a camera pair we use the same target point but add individual uniform noise to its position. The noise level depends on the distance of each camera to this target point.

We sample 34320 image pairs from the 150 annotated scenes for training. Fig. 3 shows images from this dataset and the corresponding ground truth depth maps. The generated data set contains a diverse set of scenes with many realistic images and actual ground truth data, and remedies the main disadvantages of Scenes11. However, adding new scenes to this dataset requires manual annotation, which is a time-consuming process. Due to the small number of scenes we only use this data for training. We plan to gradually extend this dataset in the future with more scenes.

**Generalization test data** In Section 6.4.1 of the main paper, we compare the generalization capabilities of DeMoN and other learned approaches. To evaluate this, we reconstructed seven self-recorded scenes using COLMAP [7, 8], and generated 16 views of these scenes with corresponding depth maps. Examples are shown in Fig. 4. Content of the scenes is very different from the data DeMoN has been trained on: Close-ups of unique objects such as a miniature bridge, figurine, and ship, as well as a person. We also include some outdoor scenes containing different architecture as well as 90-degree rotated images and a unique sculpture.

*Scale normalization:* To compensate for the inherent scale ambiguity of our reconstruction, we compute all errors in Tab. 3 of the main paper with optimally scaled depth predictions. The scale $s_{\log} = \exp(\frac{1}{n} \sum \log \hat{z} - \log z)$ is computed to minimize the mean logarithmic difference between ground truth and prediction.

## D. Experiments with Ground Truth

DeMoN iterates between estimating optical flow and estimating depth and camera motion. This is supposed to gradually refine the depth and motion estimates. The architecture is justified by a strong mathematical relation between optical flow, depth and camera motion. Optical flow can be computed analytically from depth and motion, and vice-versa, depth and motion can be computed from optical flow in non-degenerate cases. But can a convolutional network fully exploit this mathematical relation? Given perfect optical flow, can it indeed estimate depth and camera motion very accurately?

To answer these questions we trained networks to estimate depth and motion from ground truth optical flow, and, other way round, to estimate optical flow given ground truth depth and motion. Results on the SUN3D dataset are reported in Table 2. In all cases performance dramatically improves when the ground truth input is provided, compared to only taking an image pair as input. The network can use an accurate optical flow estimate to refine the depth and motion estimates, and vice-versa.

At the same time, this experiment provides an upper bound on the performance we can expect from the current architecture. DeMoN's performance is still far below this bound, meaning that the difficulties come from estimating optical flow, depth and motion from images, and not from converting between these modalities.

| Ground truth | | Depth | | | Motion | | Flow |
|---|---|---|---|---|---|---|---|
| flow | dep+mot | L1-inv | sc-inv | L1-rel | rot | tran | EPE |
| yes | no | 0.007 | 0.058 | 0.066 | - | - | - |
| yes | no | - | - | - | 0.340 | 2.235 | - |
| no | no | 0.058 | 0.163 | 0.603 | 4.472 | 41.766 | - |
| no | yes | - | - | - | - | - | 0.005 |
| no | no | - | - | - | - | - | 0.027 |

Table 2. The effect of providing the ground truth optical flow (flow) or ground truth depth and motion (dep+mot) to the network. A network can be trained to produce very accurate depth and motion estimates given the ground truth optical flow, and vice-versa, a network can estimate the optical flow very well given the ground truth depth and motion.

| Dataset | Perfect GT | Photorealistic | Outdoor scenes | Rot. avg | Rot. stddev | Tri. angle avg | Tri. angle stddev |
|---|---|---|---|---|---|---|---|
| SUN3D | no | yes | no | 10.6 | 7.5 | 5.2 | 4.6 |
| RGBD | no | yes | no | 10.4 | 8.3 | 6.8 | 4.5 |
| Scenes11 | yes | no | (yes) | 3.3 | 2.1 | 5.3 | 4.4 |
| MVS | no | yes | yes | 34.3 | 24.7 | 28.9 | 17.5 |
| Blendswap | yes | (yes) | (yes) | 23.1 | 17.1 | 20.1 | 13.6 |

Table 1. **Training dataset properties. Perfect GT:** Perfect ground truth camera poses and depth maps are only available for the synthetic datasets Scenes11 and Blendswap. **Photorealistic:** The synthetic Blendswap dataset features some photorealistic scenes, while Scenes11 looks entirely artificial. The other datasets use real images. **Outdoor scenes:** MVS is the only outdoor dataset with real images. Images from Scenes11 show wide open spaces, similar to outdoor data. Blendswap contains some outdoor scenes, but is biased towards indoor environments. **Rotation and Triangulation angle:** Camera rotation and triangulation angles are given in degree. The rotation and triangulation angle is similar for SUN3D and RGB-D SLAM. Both datasets are indoor video sequences. MVS and Blendswap also show similar characteristics, which means that the sampling procedure for Blendswap mimics the camera poses of a real outdoor dataset.

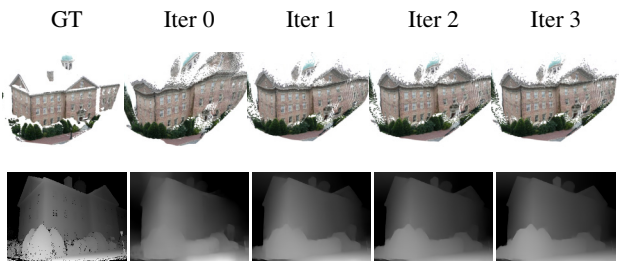

Figure 5. Effect of the iterative net on depth values.

| Iteration | Depth | | | | | | Motion | | Flow |
|---|---|---|---|---|---|---|---|---|---|
| | L1-inv | sc-inv | L1-rel | $\delta<1.25$ | $\delta<1.25^2$ | $\delta<1.25^3$ | rot | tran | EPE |
| 0 | 0.029 | 0.145 | 0.244 | 0.587 | 0.844 | 0.940 | 2.18 | 20.27 | 0.030 |
| 1 | 0.024 | 0.130 | 0.207 | 0.679 | 0.891 | 0.961 | 1.94 | 17.25 | 0.020 |
| 2 | 0.022 | 0.131 | 0.187 | 0.688 | 0.900 | 0.982 | 1.87 | 18.31 | 0.019 |
| 3 | 0.021 | 0.132 | 0.179 | 0.698 | 0.912 | 0.981 | 1.80 | 18.81 | 0.019 |
| 4 | 0.021 | 0.133 | 0.184 | 0.690 | 0.908 | 0.975 | 1.79 | 18.94 | 0.019 |
| 5 | 0.021 | 0.133 | 0.185 | 0.692 | 0.910 | 0.970 | 1.79 | 19.65 | 0.019 |

Table 3. The effect of iterative refinement of the predictions. The performance is computed on the SUN3D dataset. The results do not significantly improve beyond 3 iterations. The threshold metric is defined as the percentage of pixel $z_i$ so that $\max(\frac{z_i}{\hat{z}_i}, \frac{\hat{z}_i}{z_i}) = \delta < thr$.

## E. Effect of Iterative Refinement

The quantitative evaluation of iterative refinement is shown in Table 3. Both depth and motion accuracy significantly improve up to iteration 3.

Fig. 5 shows the effect on one sample of the MVS dataset. The iterative net improves the depth values, which is visible as a reduced distortion of the building in the point cloud.

## F. Error Distributions

We show the per-pixel error distributions and means for Base-Oracle and DeMoN on the MVS and SUN3D dataset in Fig. 6. Note that the average values in Tab. 2 in the main paper have been computed over test samples and here we average over pixels.

The distributions on the highly textured MVS show that Base-Oracle produces many very accurate depth estimates, while the distribution of errors is more spread for DeMoN. This is an effect of Base-Oracle using higher resolution images ($640 \times 480$) than DeMoN ($256 \times 192$). Base-Oracle also uses the motion ground truth, which again helps finding the correct depth.

For SUN3D distributions are more similar and the resolution adavantage of Base-Oracle is less pronounced. This can be explained by the more homogeneous image regions, which make matching difficult on this dataset. Base-Oracle suffers significantly more from outliers than DeMoN. Depending on the task higher depth accuracy can be more important than less outliers. It also shows the importance of lifting restrictions on the camera intrinsics and supporting higher resolutions in future works.

We show the distributon of the motion errors for Base-FF, Base-Mat-F and DeMoN in Fig. 7. Base-FF uses the FlowFields algorithm for correspondence search [1] and our baseline implementation using the noramlized 8-point algorithm [5] and RANSAC to compute the relative camera motion. Base-Mat-F uses the optical flow of DeMoN predicted after three iterations for correspondences and uses the Matlab implementations of [6] and RANSAC to estimate the camera motion.

The distribution is similar for Base-FF and DeMoN on the MVS dataset with Base-FF being slightly more accurate for rotation and DeMoN being more accurate for translation. Base-Mat-F is less accurate than both comparisons.

On SUN3D DeMoN can estimate the motion more accurately than the comparisons. Base-FF produces some outliers for rotation while DeMoN and Base-Mat-F have almost no outliers. Base-FF also fails to estimate accurate translation directions on SUN3D. We show some failure cases in Fig. 8. On SUN3D baselines are usually smaller than on MVS.

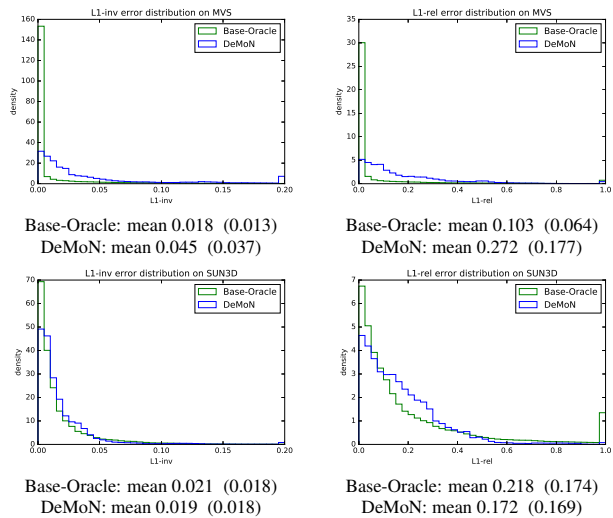

Figure 6. Error histograms and mean for per pixel depth errors (L1-inv and L1-rel) on MVS (top) and SUN3D (bottom). The last bin includes samples with errors above of the shown range. The second mean value in parenthesis excludes the last bin of the respective histogram.

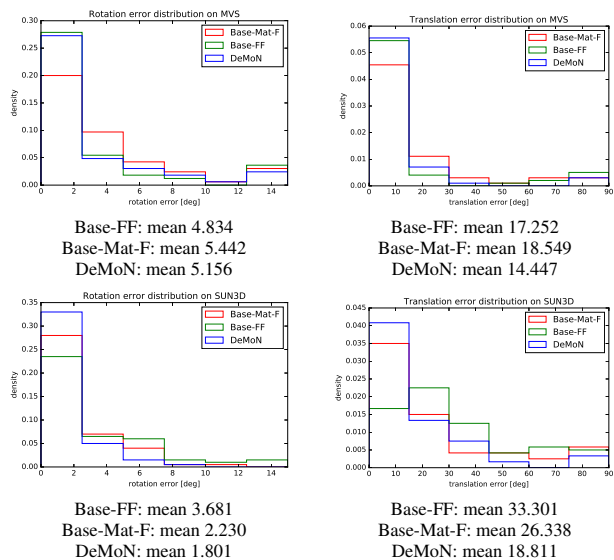

Figure 7. Error distributions for rotation and translation on MVS (top) and SUN3D (bottom). The translation error is given as angle because the length of the translation is 1 by definition. The last bin includes samples with errors above of the shown range.

## G. Qualitative Depth Estimation Results

We show more qualitative results of depth estimation on the test sets of SUN3D, RGB-D SLAM, MVS, Scenes11 and NYUv2 in Fig. 11 to Fig. 10 respectively. Our method presents smooth depth estimations while preserving sharp edges. This advantage can be observed more clearly by the

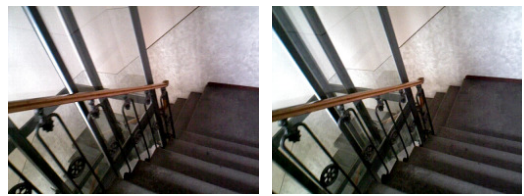

(a) DeMoN: tran 24.096, rot 0.878
Base-FF: tran 71.871, rot 2.564

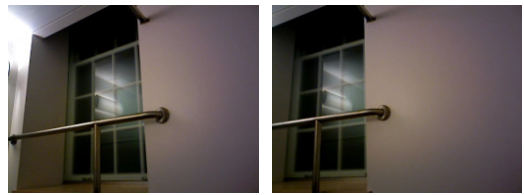

(b) DeMoN: tran 4.804, rot 1.237
Base-FF: tran 56.948 , rot 2.087

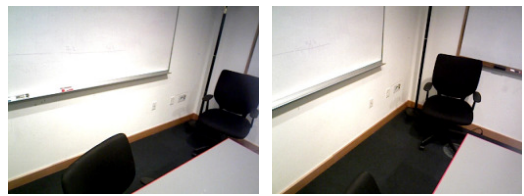

(c) DeMoN: tran 11.725, rot 1.628
Base-FF: tran 110.516, rot 15.197

Figure 8. Comparison of the motion error (in degrees) in some special cases. The classic method Base-FF fails when the baseline of the camera motion is small shown in (a) and the scenes have many homogeneous regions like (b) and (c).

point clouds, which we show in Fig. 9.

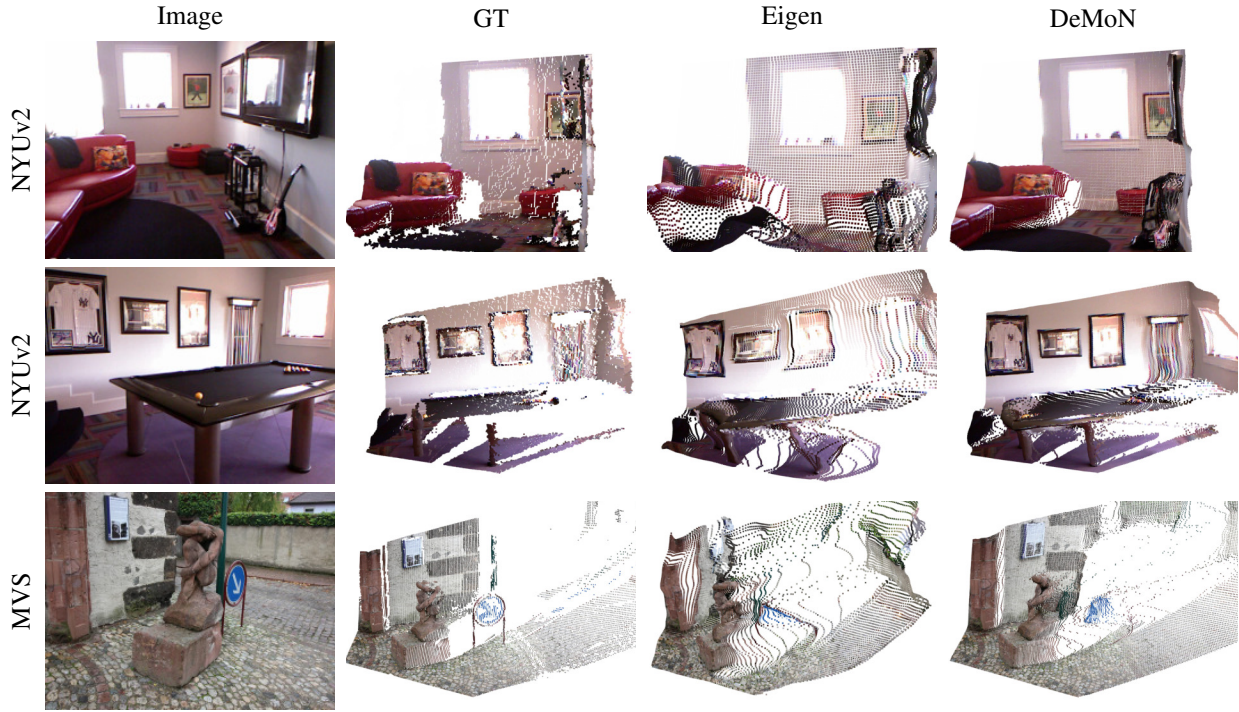

Figure 9. Qualitative point clouds comparison on NYUv2 and MVS.

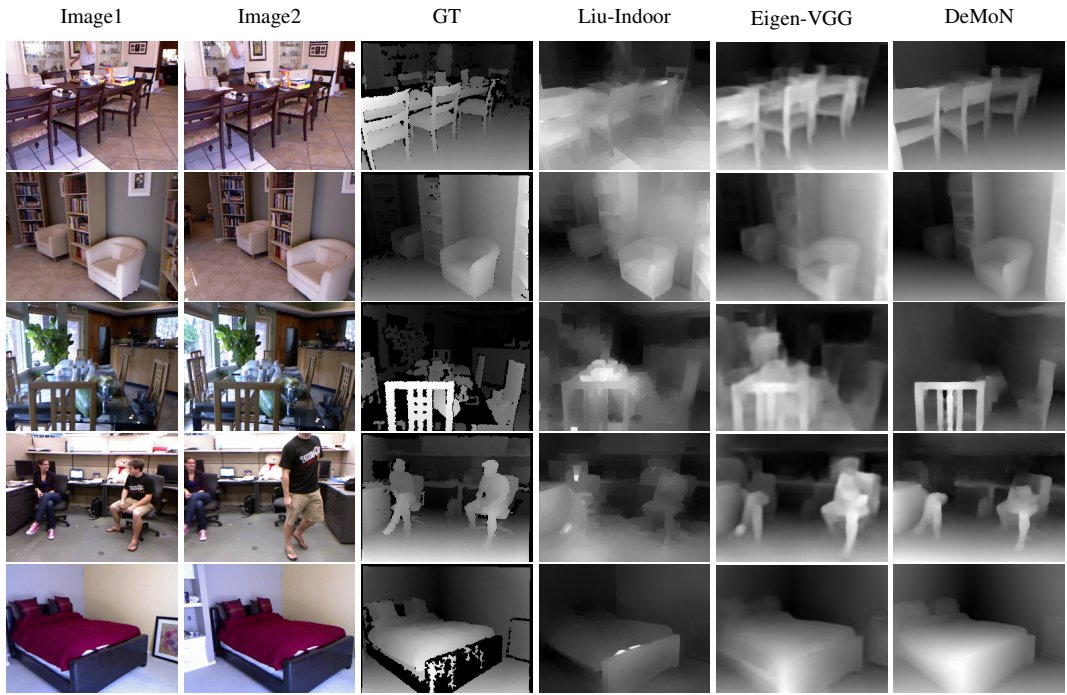

Figure 10. Qualitative depth prediction comparison on NYUv2. The Base-Oracle prediction is not available because there is no motion ground truth. Our method fails to predict the upper bodies of the persons in the fourth example because the persons move between the two frames.

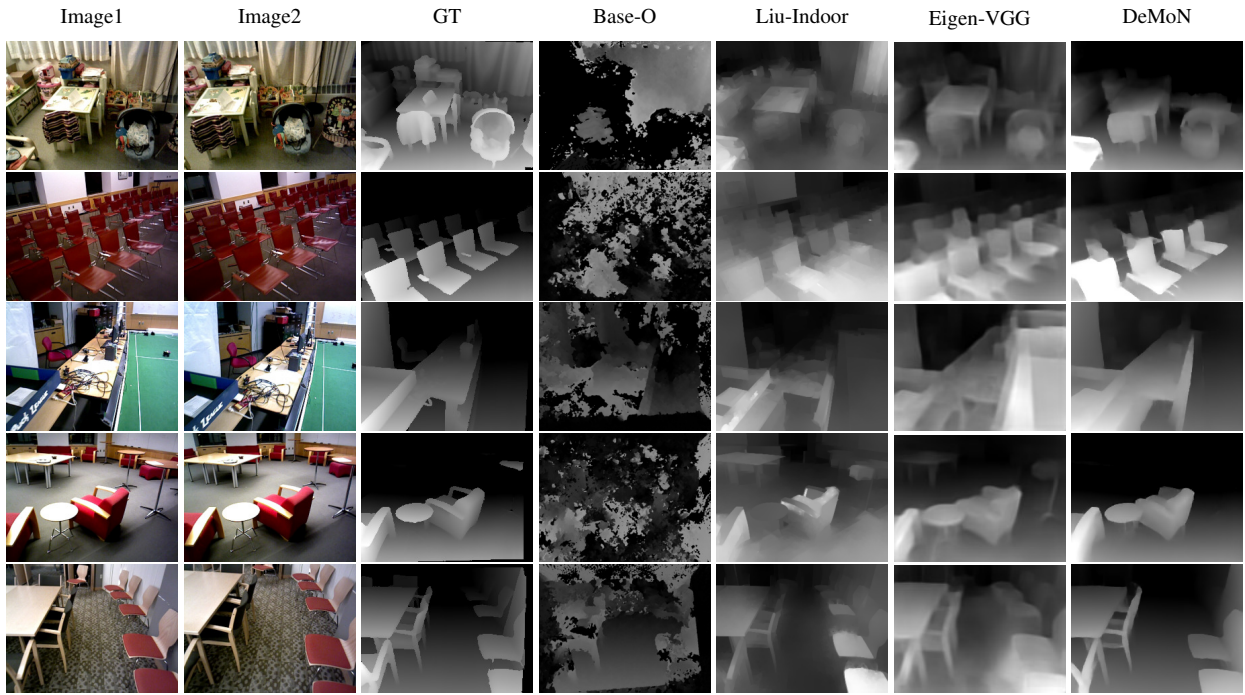

Figure 11. Qualitative depth prediction comparison on SUN3D. The Base-Oracle performs badly due to inaccurate motion ground truth. Eigen-VGG, which was trained on NYUv2, works well for many images. SUN3D is similar to the NYUv2 dataset shown in Fig. 10.

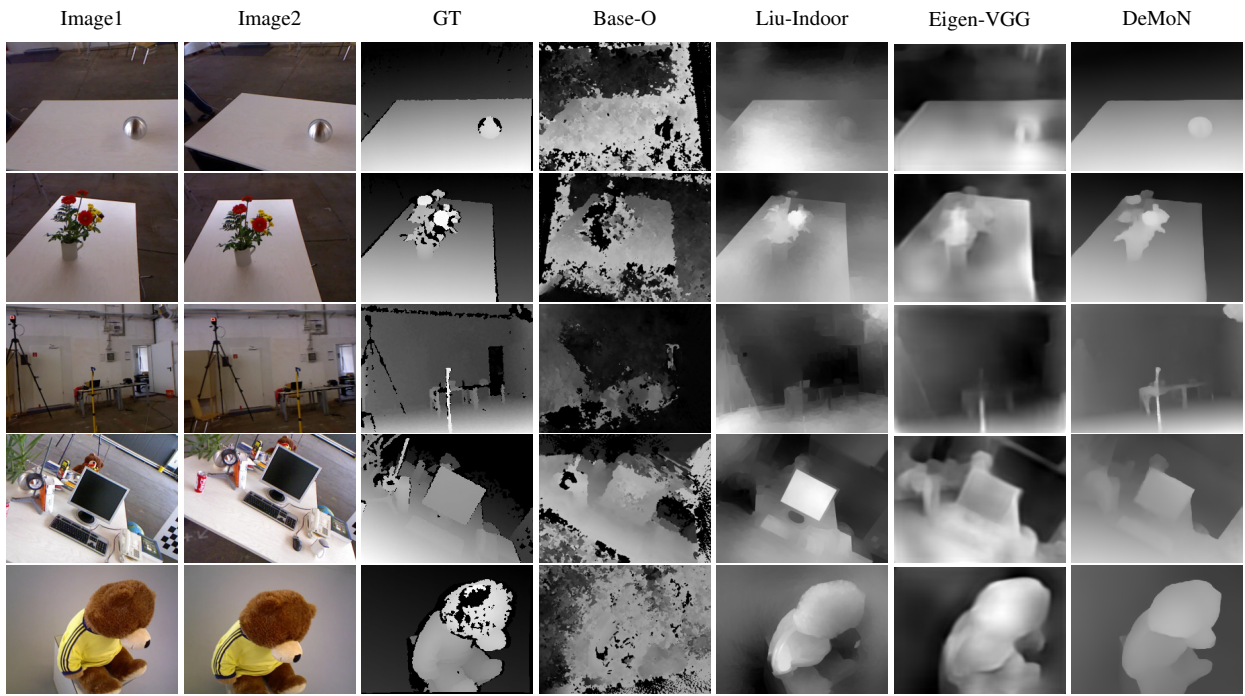

Figure 12. Qualitative depth prediction comparison on RGB-D SLAM. Our method can deal with very thin objects (third row).

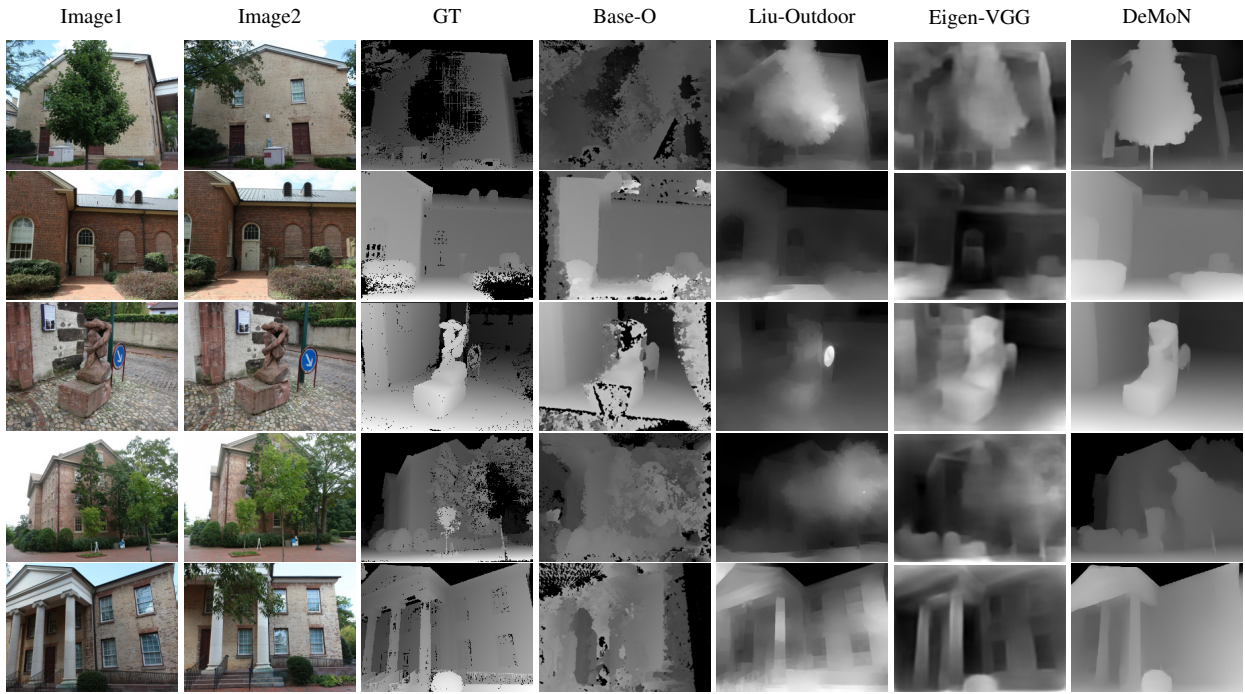

Figure 13. Qualitative depth prediction comparison on MVS. The single image methods Liu-Outdoor and Eigen-VGG do not generalize well to new datasets. The depth maps show coarse outliers caused by hallucinating wrong depth values at object contours like the street sign in the third row or the windows in the last row. The Base-Oracle method performs well on this data. Most outliers fall into image regions not visible in the second image.

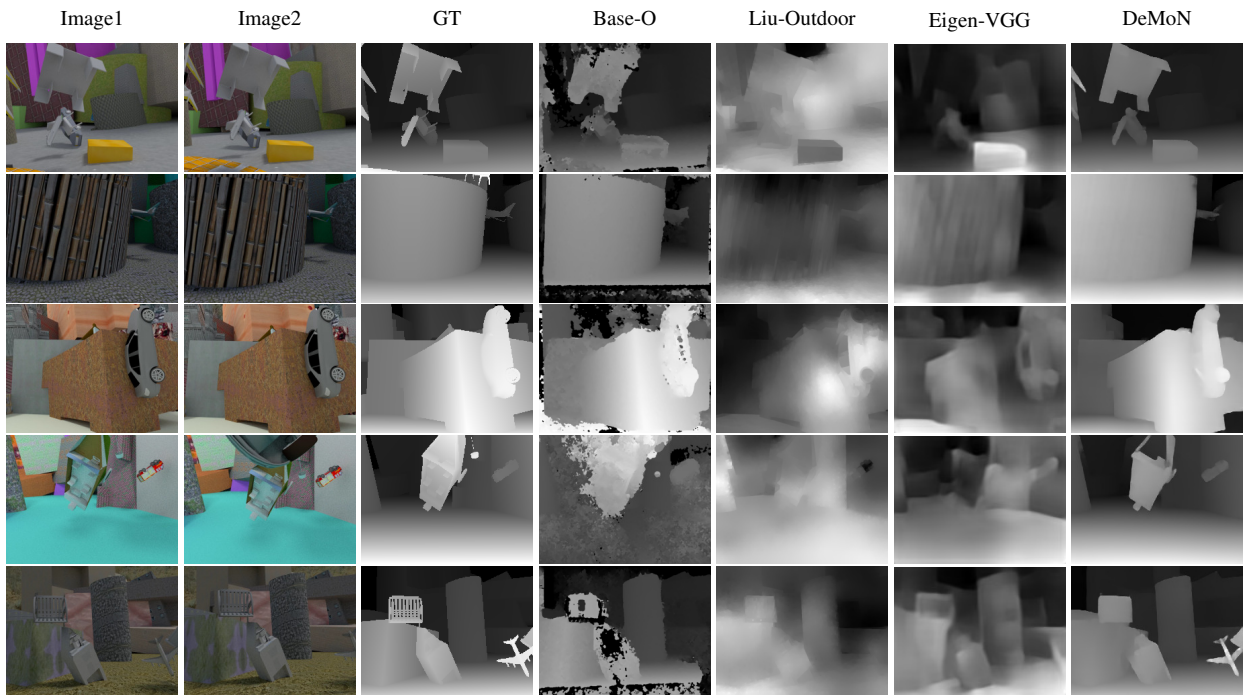

Figure 14. Qualitative depth prediction comparison on Scenes11. Base-Oracle and DeMoN give the best results on this dataset.

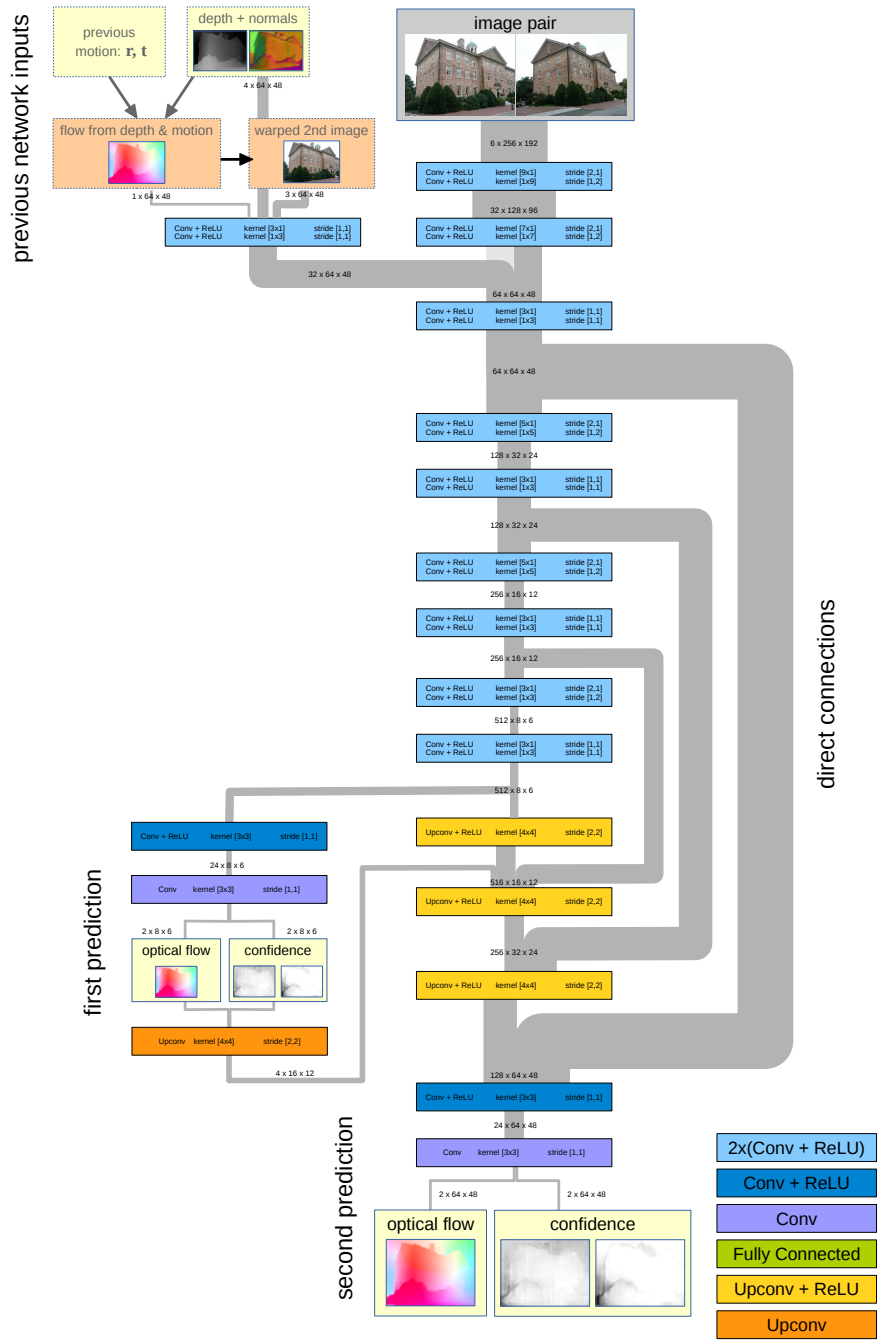

Figure 15. Encoder-decoder architecture for predicting optical flow. The depicted network is used in the bootstrap net and the iterative net part. Inputs with gray font (*depth and normals*, *flow from depth & motion*, *warped 2nd image*) are only available for the iterative net. The encoder-decoder predicts optical flow and its confidence at two different resolutions. The first prediction is directly appended to the end of the encoder and its output resolution is $8 \times 6$. We also apply our losses to this small prediction. We also feed this prediction back into the decoder part after upsampling it with an upconvolutional layer. The second prediction is part of the decoder and predicts all outputs with a resolution of $64 \times 48$, which is four times smaller than the original image dimensions ($256 \times 192$). Due to this resolution difference we concatenate inputs from the previous network at the respective spatial resolution level within the encoder. We use direct connections from the encoder to the decoder, which allows the decoder to reuse features from the respective spatial resolution levels from the encoder.

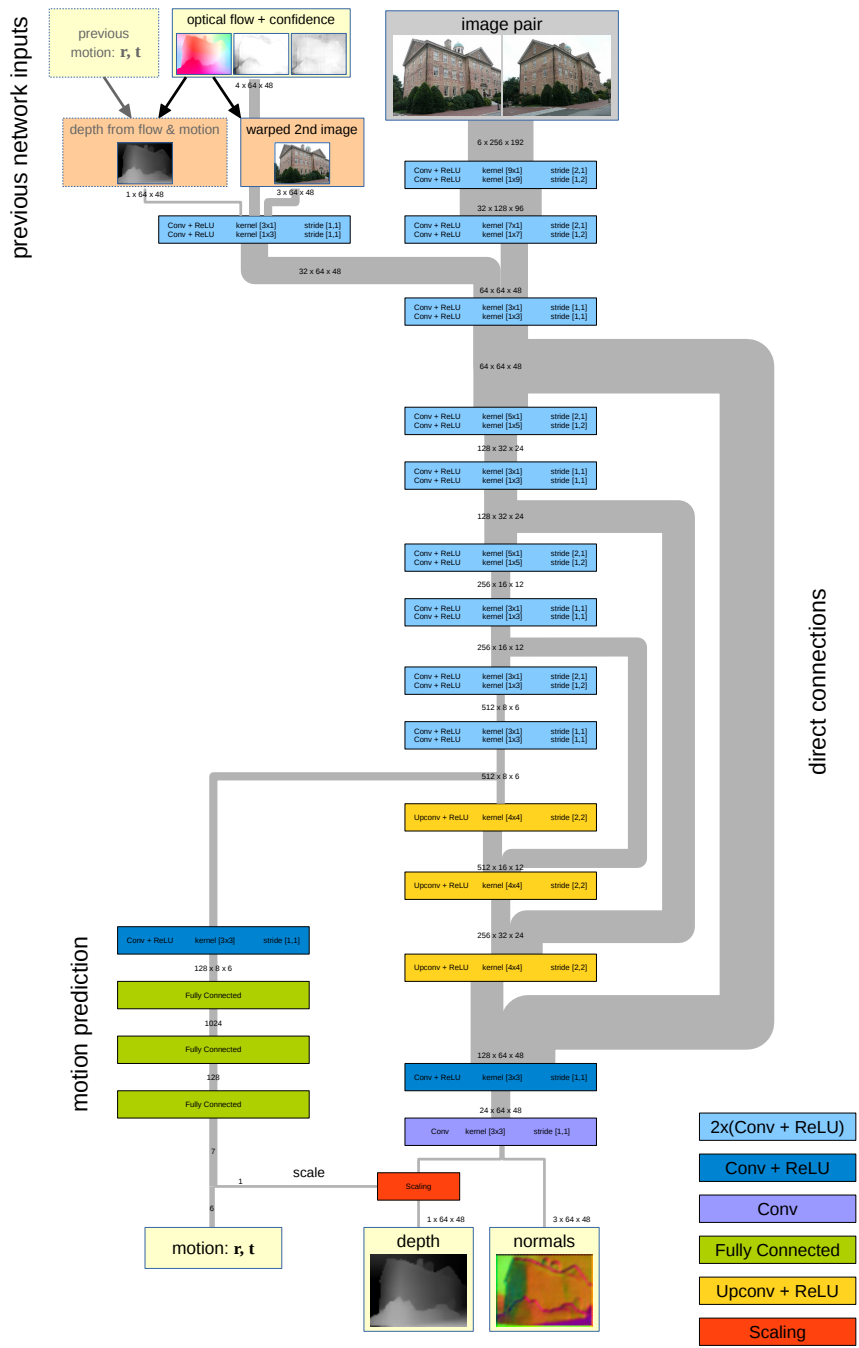

Figure 16. Encoder-decoder architecture for predicting depth and camera motion. The depicted network is used in the bootstrap net and the iterative net part. Inputs with gray font (*previous motion*, *depth from flow & motion*) are only available for the iterative net. The encoder-decoder predicts the depth map, the normals and the camera motion for an image pair. Similar to the encoder-decoder shown in Fig. 15, this encoder-decoder features direct connections and integrates previous network inputs at the corresponding resolution level into the encoder. This encoder-decoder predicts a camera motion vector and depth and normal maps. While all predictions share the encoder part, the camera motion prediction uses a separate fully connected network for its prediction. The depth and normal prediction is integrated in the decoder part. The scale of the depth values and the camera motion are highly related, therefore the motion prediction part also predicts a scale factor that we use to scale the final depth prediction.

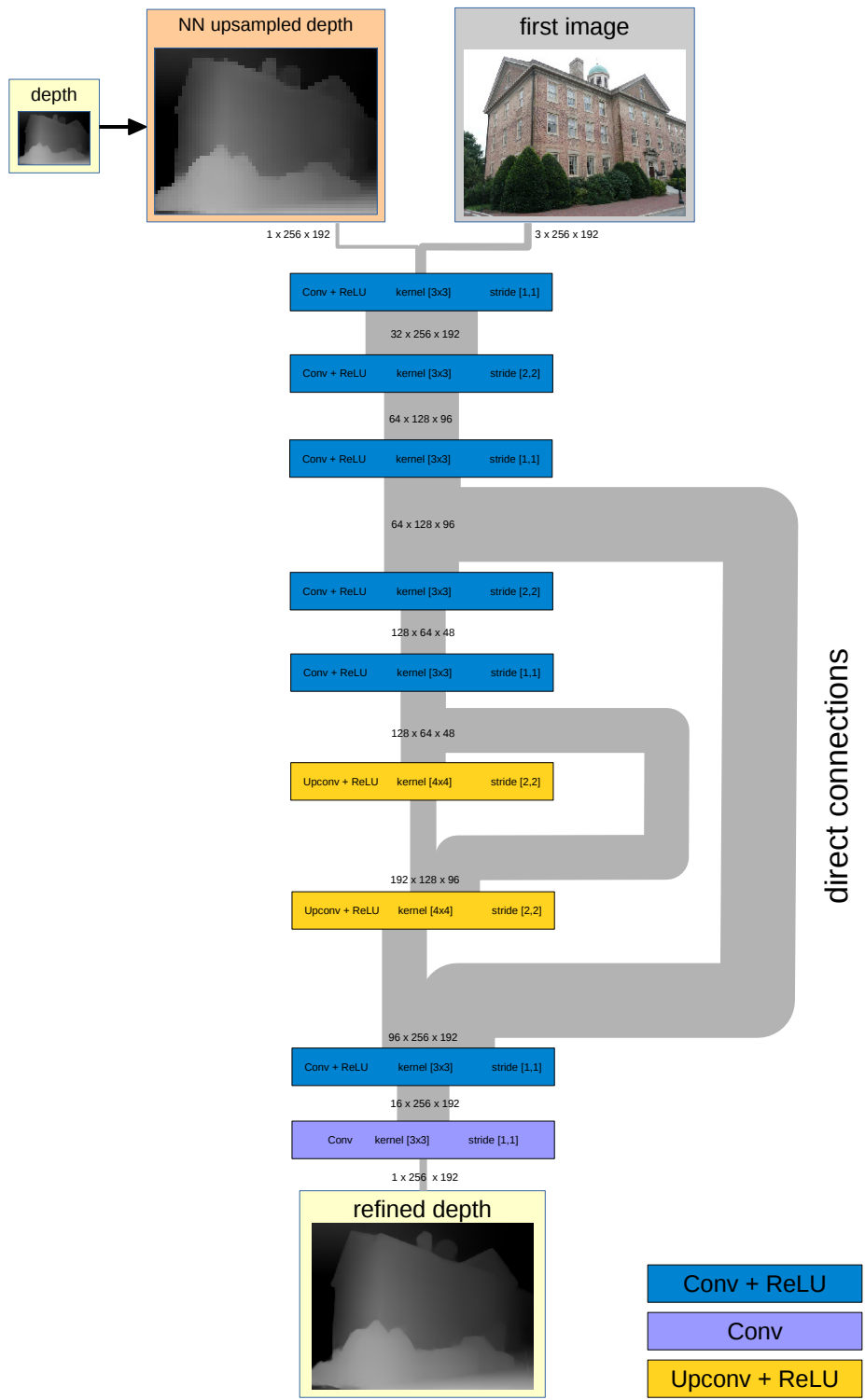

Figure 17. Encoder-decoder architecture for refining the depth prediction. The refinement network is a simple encoder-decoder network with direct connection. Input to this network is the first image and the upsampled depth map with nearest neighbor interpolation. Output is the depth map with the same resolution as the input image.

# References

[1] C. Bailer, B. Taetz, and D. Stricker. Flow Fields: Dense Correspondence Fields for Highly Accurate Large Displacement Optical Flow Estimation. In *IEEE International Conference on Computer Vision (ICCV)*, Dec. 2015. 4

[2] Blender Online Community. *Blender - a 3D modelling and rendering package*. Blender Foundation, Blender Institute, Amsterdam, 2016. 2

[3] A. X. Chang, T. Funkhouser, L. Guibas, P. Hanrahan, Q. Huang, Z. Li, S. Savarese, M. Savva, S. Song, H. Su, J. Xiao, L. Yi, and F. Yu. ShapeNet: An Information-Rich 3D Model Repository. Technical Report arXiv:1512.03012 [cs.GR], Stanford University — Princeton University — Toyota Technological Institute at Chicago, 2015. 2

[4] S. Fuhrmann, F. Langguth, and M. Goesele. Mve-a multiview reconstruction environment. In *Proceedings of the Eurographics Workshop on Graphics and Cultural Heritage (GCH)*, volume 6, page 8, 2014. 2

[5] R. I. Hartley. In defense of the eight-point algorithm. *IEEE Transactions on Pattern Analysis and Machine Intelligence*, 19(6):580–593, June 1997. 4

[6] D. Nister. An efficient solution to the five-point relative pose problem. *IEEE Transactions on Pattern Analysis and Machine Intelligence*, 26(6):756–770, June 2004. 4

[7] J. L. Schönberger and J.-M. Frahm. Structure-from-motion revisited. In *IEEE Conference on Computer Vision and Pattern Recognition (CVPR)*, 2016. 2, 3

[8] J. L. Schönberger, E. Zheng, M. Pollefeys, and J.-M. Frahm. Pixelwise view selection for unstructured multi-view stereo. In *European Conference on Computer Vision (ECCV)*, 2016. 2, 3

[9] J. Sturm, N. Engelhard, F. Endres, W. Burgard, and D. Cremers. A benchmark for the evaluation of rgb-d slam systems. In *Proc. of the International Conference on Intelligent Robot Systems (IROS)*, Oct. 2012. 2

[10] B. Ummenhofer and T. Brox. Global, dense multiscale reconstruction for a billion points. In *IEEE International Conference on Computer Vision (ICCV)*, Dec 2015. 2

[11] J. Xiao, A. Owens, and A. Torralba. SUN3D: A Database of Big Spaces Reconstructed Using SfM and Object Labels. In *IEEE International Conference on Computer Vision (ICCV)*, pages 1625–1632, Dec. 2013. 2


}
{}

\end{document}